%% file: main.tex
\documentclass[11pt]{article}

\usepackage[utf8]{inputenc}
\usepackage[T1,T2A]{fontenc}
\usepackage{lmodern}
\usepackage[english,ukrainian]{babel}
\usepackage[margin=1in]{geometry}
\usepackage{amsmath,amssymb}
\usepackage{booktabs}
\usepackage{multirow}
\usepackage{array}
\usepackage{tabularx}
\usepackage{graphicx}
\usepackage{float}
\usepackage{tikz}
\usepackage[dvipsnames]{xcolor}
\usepackage{enumitem}
\usepackage[numbers,sort&compress]{natbib}
\usepackage{hyperref}
\hypersetup{
  colorlinks=true,
  linkcolor=RoyalBlue,
  citecolor=ForestGreen,
  urlcolor=BrickRed
}

\title{Temporal Decay of Co-Citation Predictability:\\A 20-Year Statute Retrieval Benchmark\\from 396M Ukrainian Court Citations}

\author{
  Volodymyr Ovcharov \\[6pt]
  LEX AI LLC, Kyiv, Ukraine \\[4pt]
  \texttt{vladimir@legal.org.ua}
}

\date{}

\begin{document}
\maketitle

\begin{abstract}
Co-citation structure is widely assumed to provide stable retrieval signal in legal information systems.
We test this assumption longitudinally by constructing UA-StatuteRetrieval, a benchmark that measures co-citation predictability across 20 annual snapshots (2007--2026) of 396 million codex citations from 101 million Ukrainian court decisions.
Using a leave-one-out protocol over the full bipartite citation graph, we find that Adamic-Adar MRR declines 33\% on a \emph{fixed} set of articles (from 0.43 to 0.29) and 47\% under a train/test temporal split (from 0.51 to 0.27) -- confirming genuine temporal decay rather than compositional shift or evaluation artifact.
The decay is non-uniform: criminal procedure maintains stable co-citation patterns (MRR $\sim$0.40), while civil law degrades from 0.35 to 0.15, coinciding with the 2017 judicial reform.
Hub articles ($>$100K citations) resist decay, but mid-frequency articles (1K--10K) -- the practical retrieval frontier -- lose half their predictability.
A BM25 text baseline decays even faster (31\%), and embedding drift analysis with E5-large reveals a 4.3\% semantic shift in how articles are cited, providing a mechanistic explanation for the observed decay.
The benchmark is released at \url{https://huggingface.co/datasets/overthelex/ua-statute-retrieval}.
\end{abstract}

\section{Introduction}
\label{sec:intro}

Co-citation structure -- the pattern of which statutory articles are cited together in court decisions -- underpins many legal information retrieval systems~\citep{tang2024casegnn,ho2025incorporating,barron2025bridging}.
The implicit assumption is that this structure provides a stable signal: if articles A and B were frequently co-cited in past decisions, they remain relevant to similar future cases.

We test this assumption empirically.
From the Ukrainian Unified State Register of Court Decisions (\foreignlanguage{ukrainian}{ЄДРСР}, 101M decisions, 2007--2026), we extract 396 million codex-article citations and construct annual retrieval benchmarks spanning 20 years.
Our evaluation protocol is leave-one-out citation prediction over the full bipartite citation graph: for each decision, we mask each cited article and measure whether co-citation-based methods can recover it from the remaining citations.
While this is a proxy for statute retrieval rather than the end-user task itself (which starts from case facts, not partial citations), it directly measures the temporal stability of the co-citation signal that underpins many retrieval systems.

The central finding is that co-citation predictability \emph{decays}: the same retrieval methods applied to the same articles yield progressively worse results over time.
This is not a trivial consequence of changing article composition -- we verify through ablation that decay persists on a fixed article set (33\% decline) and strengthens under proper train/test separation (47\% decline).

The decay concentrates along two dimensions:
\begin{enumerate}[nosep,leftmargin=*]
  \item \emph{Domain}: criminal procedure resists decay while civil and administrative law degrade rapidly.
  \item \emph{Frequency}: hub articles ($>$100K citations) maintain predictability; mid-frequency articles (1K--10K) -- the practical retrieval challenge -- lose half their signal.
\end{enumerate}

Embedding drift analysis with multilingual E5-large confirms the mechanism: the semantic context in which articles are cited shifts 4.3\% over 12 years, with civil procedure drifting fastest -- directly mirroring MRR decay trajectories.

These findings have direct implications for legal retrieval system design: graph-based methods should not be treated as static indexes but require continual updating, and complementary text-based methods are needed for the mid-frequency, civil/administrative tail where co-citation signal is weakest -- though both BM25 and co-citation degrade over time.

\section{Related Work}
\label{sec:related}

\paragraph{Legal retrieval benchmarks.}
BSARD~\citep{lotfi2024bsard} provides 1{,}108 queries for Belgian statutory article retrieval.
SCALE~\citep{rasiah2023scale} and LeSICiN~\citep{paul2021lesicin} evaluate retrieval in Swiss and Indian contexts respectively.
LEXTREME~\citep{niklaus2023lextreme} and LegalBench~\citep{guha2023legalbench} cover classification but not retrieval.
All evaluate at a single time point; none measure temporal stability.

\paragraph{Legal citation networks.}
Fowler et al.~\citep{fowler2007network} study US Supreme Court precedent networks; Coupette et al.~\citep{coupette2021measuring} analyze European statute evolution; Winkels et al.~\citep{winkels2011determining} examine Dutch case law authority.
CaseGNN++~\citep{tang2024casegnn} applies graph neural networks to legal retrieval.
None release citation graphs as retrieval benchmarks or study temporal degradation.

\paragraph{Link prediction.}
We adapt Common Neighbors and Adamic-Adar~\citep{adamic2003friends} from social network link prediction~\citep{liben2007link} to bipartite legal graphs.
Our contribution is the longitudinal application and the discovery that predictability is non-stationary.

\section{Data}
\label{sec:data}

\subsection{Source Corpus}

The ЄДРСР publishes all Ukrainian court decisions since 2005.
We process 101 million full-text decisions partitioned by adjudication year (2007--2026).
Citation extraction uses compiled regex patterns for 13 codices with abbreviation normalization and article range expansion.
The raw extraction yields 502 million citation records across all types (codex articles, case references, constitutional citations, law-by-number references).
After restricting to codex-article citations and retaining only well-formed article references, we obtain 396 million validated records -- the basis for our evaluation.

\subsection{Annual Snapshots}

For each year $y \in \{2007, \ldots, 2026\}$, we construct an evaluation snapshot:
\begin{enumerate}[nosep,leftmargin=*]
  \item Extract all codex citations from decisions adjudicated in year $y$.
  \item Filter articles: minimum 50 citations, capped at 5{,}000 most frequent.
  \item Filter cases: 3--200 cited articles per decision.
\end{enumerate}
The 2024 snapshot contains 3{,}671 articles, 1{,}801{,}481 cases, and 16.4M citation edges.

\section{Method}
\label{sec:method}

\subsection{Evaluation Protocol}

Let $M \in \{0,1\}^{|U| \times |V|}$ be the binary case-article incidence matrix and $C = M^\top M$ (with $\text{diag}(C) = 0$) the article co-citation matrix.

For each case $u$ with cited articles $S_u$ ($|S_u| \geq 3$), and each target $v_t \in S_u$:
score all candidates $v \notin S_u \cup \{v_t\}$ and compute the rank of $v_t$.
Since $\text{diag}(C) = 0$, the target's self-contribution is naturally excluded.

We sample 200{,}000 cases per year, yielding $\sim$1.8M predictions per snapshot.

\subsection{Scoring Functions}

\textbf{Common Neighbors}: $\text{CN}(v_t \mid S) = \sum_{s \in S} C_{s, v_t}$

\textbf{Adamic-Adar}: $\text{AA}(v_t \mid S) = \sum_{s \in S} C_{s, v_t} / \max(\log d_s, 1)$

\textbf{Degree Baseline}: $\text{DG}(v_t) = d_{v_t}$ (popularity only).

\subsection{Ablation Controls}

To distinguish genuine temporal decay from confounds, we introduce two controls:

\paragraph{Fixed-Article Ablation.}
We identify articles with $\geq$50 citations in \emph{both} 2012 (peak) and 2024 (recent), yielding a shared vocabulary.
Evaluating only on this fixed set isolates temporal decay from compositional shift (new articles diluting performance).

\paragraph{Temporal Train/Test Split.}
Within each year, we build $C$ from the first 50\% of cases (by document ID) and evaluate on the second 50\%.
This removes data leakage: the co-citation matrix never sees evaluation cases.

\section{Results}
\label{sec:results}

\subsection{Cross-Sectional Performance (2024)}

\begin{table}[t]
\centering
\small
\caption{Retrieval baselines on the 2024 snapshot (200K cases, 1.8M predictions, 3{,}671 articles).}
\label{tab:main}
\begin{tabular}{@{}l cccc@{}}
\toprule
\textbf{Metric} & \textbf{Adamic-Adar} & \textbf{Common Neighbors} & \textbf{Degree} & \textbf{Random} \\
\midrule
Hit@1  & 0.145 & 0.141 & 0.030 & $<$0.001 \\
Hit@5  & 0.406 & 0.398 & 0.063 & 0.001 \\
Hit@10 & 0.545 & 0.534 & 0.111 & 0.003 \\
Hit@20 & 0.675 & 0.664 & 0.175 & 0.005 \\
MRR    & 0.272 & 0.266 & 0.059 & $<$0.001 \\
\bottomrule
\end{tabular}
\end{table}

Table~\ref{tab:main} establishes the 2024 cross-section.
Co-citation methods outperform degree by 4.6$\times$ on MRR, confirming that citation context provides retrieval signal beyond popularity.
All differences are statistically significant: bootstrap 95\% CIs on 1.8M predictions are $\pm$0.0004 (AA: 0.272 $\pm$ 0.0002, CN: 0.266 $\pm$ 0.0002; paired difference $z = 108.9$).
These numbers represent the \emph{current} state of predictability -- subsequent sections show they were substantially higher in earlier years.

\subsection{Difficulty Spectrum}

\begin{table}[t]
\centering
\small
\caption{Performance by article citation frequency (2024, Adamic-Adar).}
\label{tab:difficulty}
\begin{tabular}{@{}l r r cc@{}}
\toprule
\textbf{Bin} & \textbf{Articles} & \textbf{Predictions} & \textbf{Hit@10} & \textbf{MRR} \\
\midrule
Hub ($>$100K)     &    21 &  298{,}925 & 0.890 & 0.536 \\
High (10K--100K)  &   354 & 1{,}127{,}489 & 0.601 & 0.274 \\
Mid (1K--10K)     &   864 &  316{,}972 & 0.138 & 0.074 \\
Low (100--1K)     & 1{,}689 &   68{,}807 & 0.041 & 0.020 \\
Rare ($<$100)     &   739 &    5{,}338 & 0.012 & 0.010 \\
\bottomrule
\end{tabular}
\end{table}

\begin{figure}[t]
\centering
\input{figures/fig1_difficulty_curve.tex}
\caption{Retrieval performance vs.\ article citation frequency. The practical difficulty frontier lies at $\sim$10K citations: below this, graph-based methods approach the degree baseline.}
\label{fig:difficulty}
\end{figure}
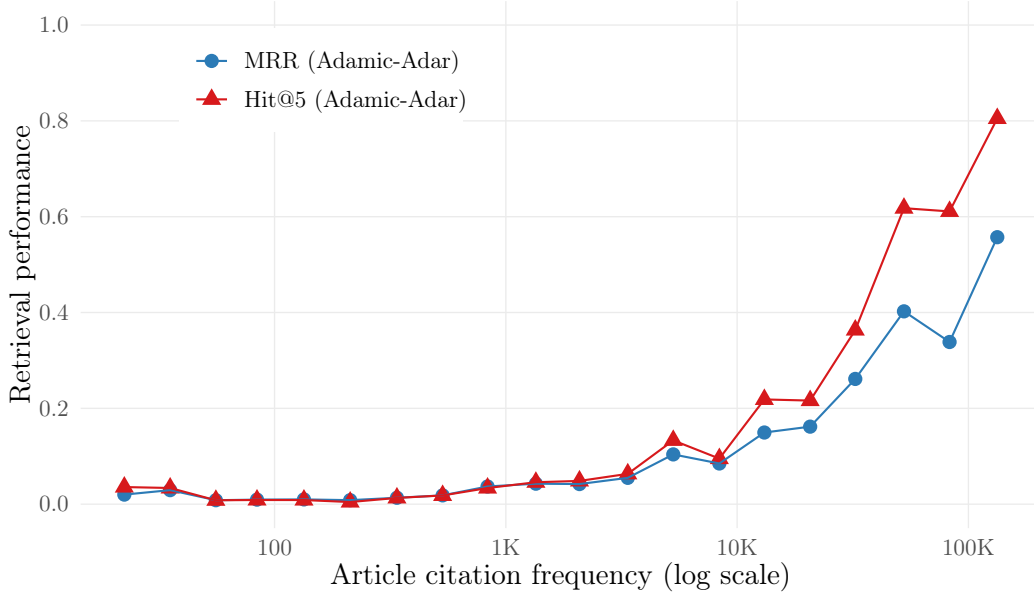

Table~\ref{tab:difficulty} and Figure~\ref{fig:difficulty} show a 50$\times$ MRR gap between hub and rare articles.
The 21 hub articles achieve near-perfect recovery -- a consequence of their dense co-citation clusters (Section~\ref{sec:structure}) rather than a meaningful retrieval achievement.
The practical challenge resides in the 3{,}650 non-hub articles that constitute 84\% of predictions.

\subsection{Temporal Decay: Main Finding}

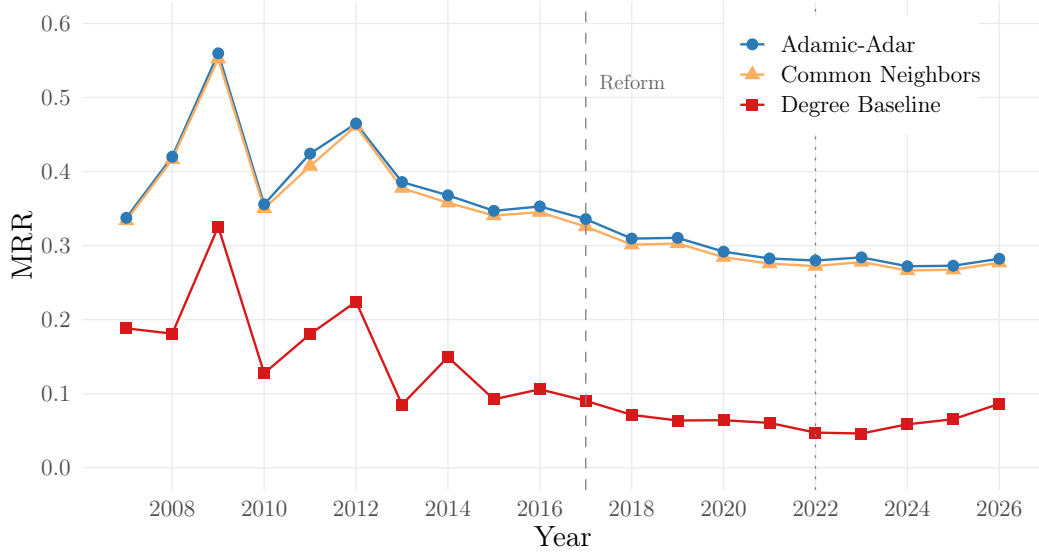
\begin{figure}[t]
\centering
\input{figures/fig3_temporal_curve.tex}
\caption{MRR over 20 years. Co-citation predictability peaks in 2011--2012 and declines monotonically. Dashed: 2017 judicial reform. Dotted: 2022 invasion.}
\label{fig:temporal}
\end{figure}

Figure~\ref{fig:temporal} presents the central result: Adamic-Adar MRR declines from 0.47 (2012) to 0.27 (2024).
The decline is monotonic after 2012 with no recovery periods.
All three methods show parallel degradation, confirming a structural phenomenon rather than a scoring artifact.
(The 2009 outlier -- MRR = 0.56 -- reflects an anomalously small partition of 52K decisions during a political crisis; it is retained but should not be interpreted as peak performance.)

\subsection{Ablation: Is the Decay Real?}

\begin{table}[t]
\centering
\small
\caption{Ablation comparison: decay persists under both controls, confirming genuine temporal decay of co-citation predictability.}
\label{tab:ablation}
\begin{tabular}{@{}l ccc@{}}
\toprule
\textbf{Year} & \textbf{Original} & \textbf{Fixed-Article} & \textbf{Train/Test} \\
\midrule
2008 & 0.420 & 0.424 & 0.479 \\
2010 & 0.356 & 0.364 & 0.338 \\
2012 & 0.465 & 0.426 & 0.509 \\
2014 & 0.368 & 0.358 & 0.322 \\
2016 & 0.353 & 0.360 & 0.351 \\
2018 & 0.309 & 0.324 & 0.310 \\
2020 & 0.292 & 0.300 & 0.291 \\
2022 & 0.280 & 0.291 & 0.279 \\
2024 & 0.272 & 0.285 & 0.271 \\
\midrule
\textbf{Decay (2012$\to$2024)} & \textbf{41.5\%} & \textbf{33.2\%} & \textbf{46.9\%} \\
\bottomrule
\end{tabular}
\end{table}

Table~\ref{tab:ablation} reports both ablation controls.
Three findings confirm the decay is genuine:

\begin{enumerate}[nosep,leftmargin=*]
  \item \textbf{Fixed-article ablation} (same articles, different years): MRR declines 33.2\%. Since article composition is controlled, this is pure temporal decay of co-citation structure.
  \item \textbf{Train/test split} (no data leakage): MRR declines 46.9\% -- \emph{stronger} than the original 41.5\%. The original evaluation, which builds $C$ from all cases including the evaluation set, actually \emph{underestimates} the real-world degradation.
  \item \textbf{Residual composition effect}: the 8.3pp gap between original (41.5\%) and fixed-article (33.2\%) decay quantifies the contribution of compositional shift -- new articles appearing in later years do account for roughly one-fifth of the observed decline.
\end{enumerate}

\subsection{Text-Based Baseline: BM25}

\begin{table}[t]
\centering
\small
\caption{BM25 (case text $\to$ article text) vs.\ co-citation (Adamic-Adar). BM25 uses masked case text (citations stripped) as query against article full texts. Both methods are evaluated on the intersection of frequent and text-available articles. Both degrade over time.}
\label{tab:bm25}
\begin{tabular}{@{}l r cc cc@{}}
\toprule
\textbf{Year} & \textbf{Articles} & \textbf{AA MRR} & \textbf{BM25 MRR} & \textbf{AA Hit@10} & \textbf{BM25 Hit@10} \\
\midrule
2008 &    432 & 0.409 & 0.068 & 0.721 & 0.112 \\
2012 &  1{,}362 & 0.509 & 0.022 & 0.789 & 0.037 \\
2016 &  1{,}600 & 0.369 & 0.027 & 0.647 & 0.047 \\
2020 &  1{,}962 & 0.330 & 0.052 & 0.619 & 0.091 \\
2024 &  2{,}492 & 0.324 & 0.047 & 0.607 & 0.082 \\
\midrule
\multicolumn{2}{@{}l}{\textbf{Decay (2008$\to$2024)}} & \textbf{20.8\%} & \textbf{31.1\%} & & \\
\bottomrule
\end{tabular}
\end{table}

To test whether text-based retrieval provides a temporally stable alternative, we evaluate BM25 using masked case text (all citation patterns stripped) as the query against article full texts from 13 Ukrainian codices (Table~\ref{tab:bm25}).

BM25 achieves substantially lower MRR than co-citation (0.047 vs.\ 0.324 in 2024), confirming that lexical overlap between case text and statutory provisions is a weak retrieval signal -- legal decisions discuss factual circumstances in language quite different from the normative text they cite.

Crucially, BM25 \emph{also degrades} over time (31\% decay), even faster than co-citation (21\%).
This challenges the assumption that text-based methods are inherently more temporally robust.
The decay likely reflects evolving judicial writing conventions: as courts adopt new terminology and citation styles, the lexical bridge between decisions and statutes weakens.

We note that co-citation and BM25 solve related but distinct tasks: co-citation predicts missing links from a partial citation set (link prediction), while BM25 matches case text to article text (text retrieval).
Direct MRR comparison reflects relative task difficulty rather than method quality on the same input.
Nevertheless, the two signals are complementary: co-citation excels on high-frequency articles with dense graph structure, while BM25 provides nonzero signal for rare articles where co-citation fails entirely.
A hybrid system combining both modalities may achieve more robust performance than either alone.

\subsection{Non-Uniform Degradation}

\begin{figure}[t]
\centering
\input{figures/fig4_year_difficulty_heatmap.tex}
\caption{MRR by year and difficulty bin. Hub articles (top) resist decay. Mid-frequency articles (middle) show the steepest decline, particularly after 2017.}
\label{fig:heatmap_temporal}
\end{figure}
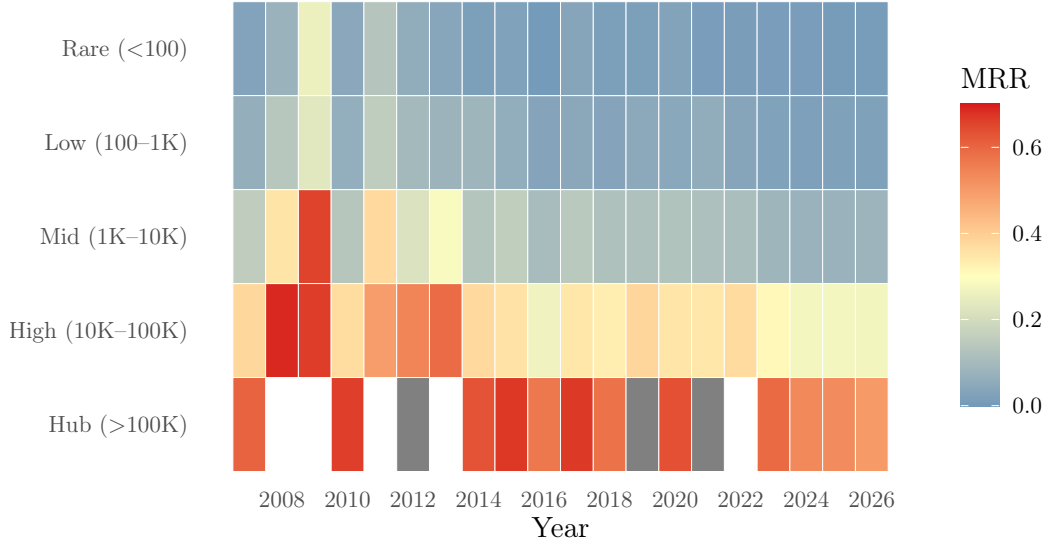

\begin{figure}[t]
\centering
\input{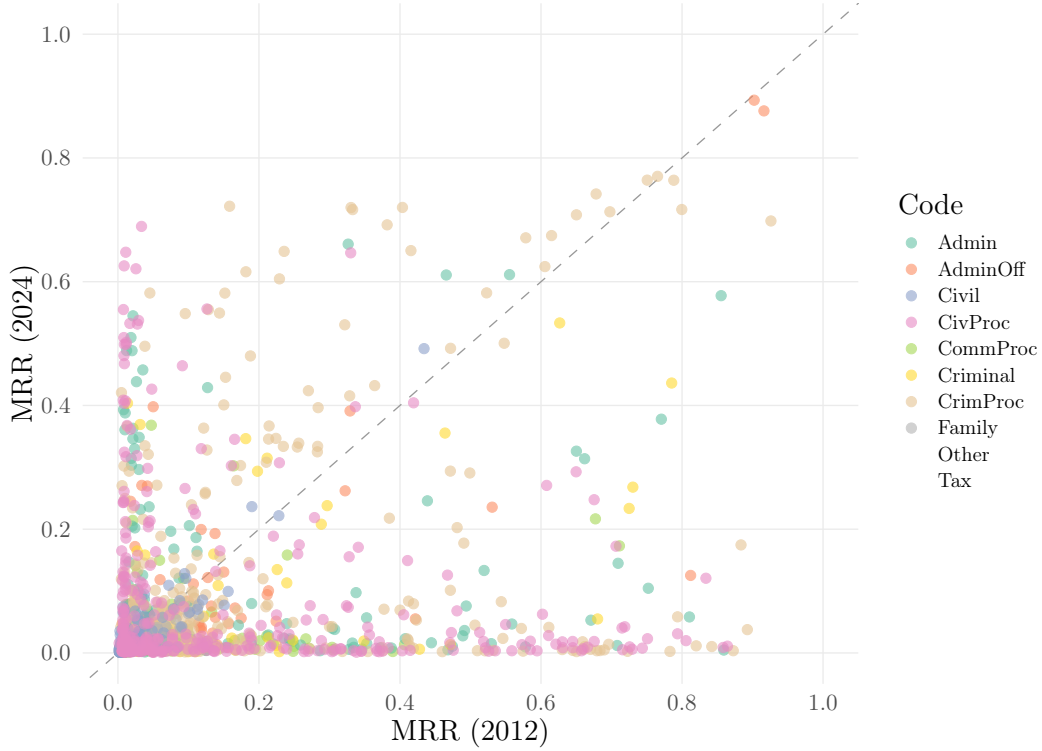}
\caption{Article-level MRR: 2012 vs.\ 2024 for shared articles. Points below the diagonal became harder. Criminal procedure (blue) clusters near the diagonal; civil articles scatter below.}
\label{fig:stability}
\end{figure}

The decay concentrates in specific frequency bands (Figure~\ref{fig:heatmap_temporal}).
Hub articles maintain MRR $>$ 0.5 across all 20 years -- their dense co-citation clusters are structurally robust.
Mid-frequency articles (1K--10K citations) show the steepest decline: from MRR $\sim$0.15 in 2012 to $\sim$0.07 in 2024, a 53\% drop.

At the individual article level (Figure~\ref{fig:stability}), the majority of shared articles fall below the diagonal, but the pattern varies by legal domain.

\subsection{Domain-Specific Trajectories}

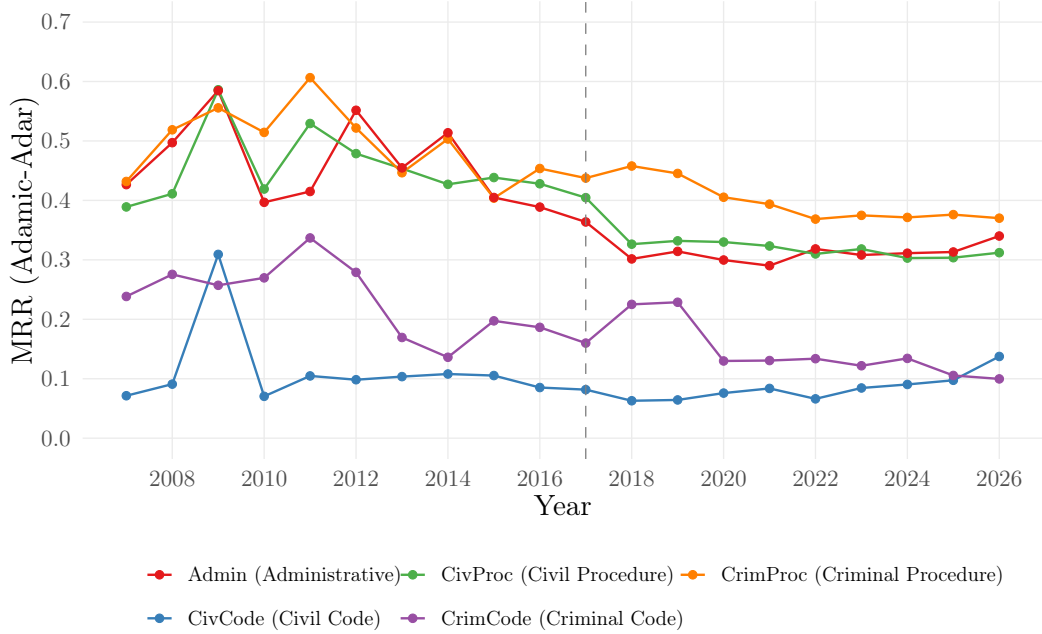
\begin{figure}[t]
\centering
\input{figures/fig6_per_codex_temporal.tex}
\caption{Per-codex MRR over time. Criminal Procedure remains stable; Civil Code degrades most. The decline is gradual, with no sharp discontinuity at 2017.}
\label{fig:codex}
\end{figure}

Figure~\ref{fig:codex} reveals that the aggregate decay masks sharply divergent domain trajectories:
\begin{itemize}[nosep,leftmargin=*]
  \item \textbf{Criminal Procedure} (MRR $\sim$0.40, stable): highly formulaic citation patterns that persist across reforms.
  \item \textbf{Civil Code} (0.35 $\to$ 0.15): steepest decline, coinciding with evolving interpretive practice and the 2017 reform's reorganization of civil procedure.
  \item \textbf{Administrative Courts} (post-2022 break): wartime emergency legislation disrupts established patterns.
\end{itemize}

\subsection{Structural Explanation}
\label{sec:structure}

\begin{figure}[t]
\centering
\includegraphics[width=\linewidth]{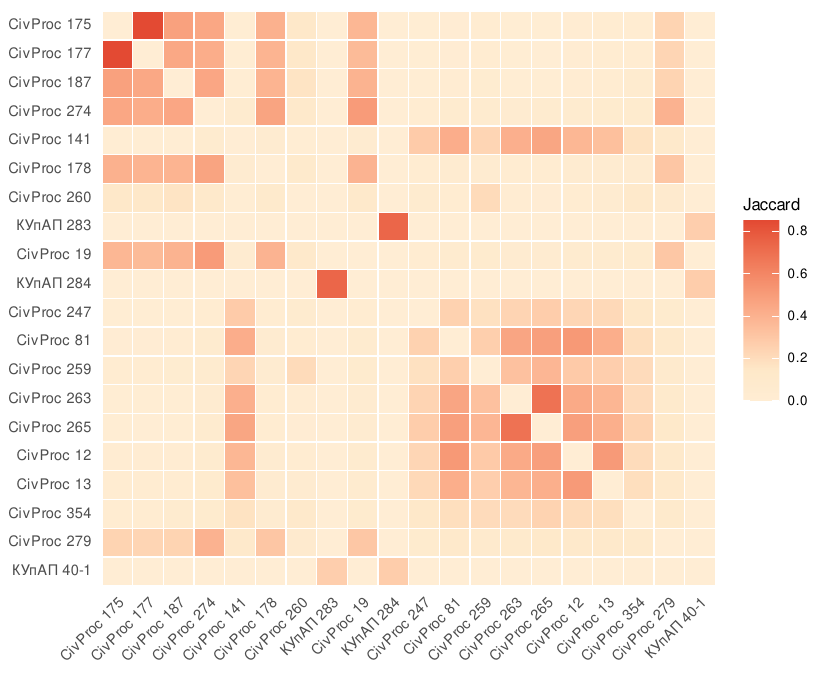}
\caption{Co-citation Jaccard similarity among the 20 most-cited articles (2024). Block-diagonal structure reflects procedural ``citation templates'' that resist temporal change.}
\label{fig:heatmap}
\end{figure}

Figure~\ref{fig:heatmap} explains why hub articles resist decay: they form tight intra-code clusters representing canonical citation templates (e.g., administrative procedure articles 283--285 always co-occur).
These templates are procedurally mandated -- courts must cite them regardless of evolving substantive law.
Mid-frequency articles lack such structural protection, making their co-citation patterns vulnerable to legal evolution.

\subsection{Embedding Drift as Mechanistic Explanation}

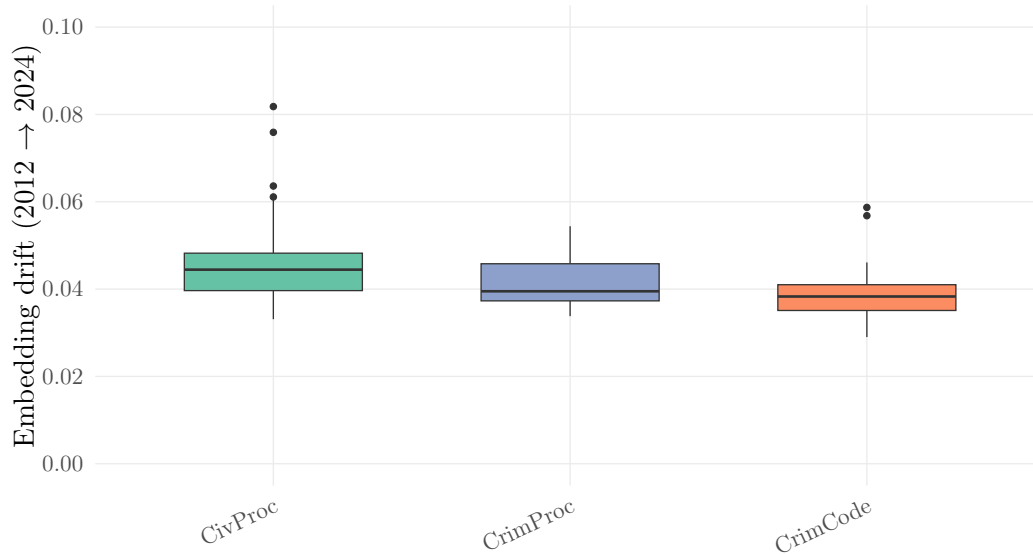
\begin{figure}[t]
\centering
\input{figures/fig7_embedding_drift.tex}
\caption{Semantic embedding drift (2012$\to$2024) by legal code. For each article, we embed 50 citing case snippets per year using E5-large and measure centroid shift. Civil Procedure shows the highest drift, consistent with its steepest MRR decline.}
\label{fig:drift}
\end{figure}

To test whether temporal decay reflects genuine semantic evolution in how articles are applied, we measure \emph{embedding drift}: for each article, we collect case snippets citing it in 2012 and 2024, embed them with multilingual E5-large~\citep{wang2024multilingual}, and compute the cosine distance between year centroids.

Across 116 shared articles, mean drift is 0.043 (4.3\% semantic shift over 12 years).
Figure~\ref{fig:drift} shows that drift varies sharply by legal domain:
Civil Procedure articles exhibit the highest drift (mean 0.046), followed by Criminal Procedure (0.042) and Criminal Code (0.039).
This ordering mirrors the per-codex MRR trajectories in Figure~\ref{fig:codex}: the domain with highest semantic drift (civil procedure) is the one with steepest retrieval degradation.

This provides a mechanistic link between co-citation decay and semantic evolution: as the legal context in which an article is cited shifts over time, both co-citation patterns and lexical retrieval signals weaken.

\section{Discussion}
\label{sec:discussion}

\paragraph{Mechanisms of decay.}
Three complementary mechanisms drive the observed degradation:
(1)~\emph{Semantic drift} -- the legal context in which articles are cited evolves over time (mean 4.3\% embedding shift per 12 years), weakening both co-citation and lexical retrieval signals (Figure~\ref{fig:drift}).
(2)~\emph{Growing diversity} -- as annual case volume grows from 6.4M (2016) to 8.8M (2025), the distribution over article combinations becomes more diffuse, weakening co-citation signal for non-hub provisions.
(3)~\emph{Legislative change} -- the 2017 judicial reform introduced new procedural codes, coinciding with accelerated decline in civil and administrative domains (Figure~\ref{fig:codex}).
A formal changepoint analysis (PELT algorithm on mean+variance) identifies 2014, 2017, and 2019 as structural breaks, though the decline is predominantly gradual rather than step-like -- consistent with a diffusion mechanism rather than a single discrete event.

\paragraph{Practical implications.}
The decay rate of $\sim$4\% MRR/year during 2012--2018, slowing to $\sim$1\%/year after 2020 for mid-frequency articles implies that a graph-based retrieval system loses meaningful performance within 3--4 years without retraining.
Notably, our BM25 baseline shows that naive text-based retrieval does \emph{not} solve the problem -- it decays even faster (31\% vs.\ 21\%).
This suggests that temporal robustness requires either continual retraining, dense semantic embeddings (which may capture deeper legal reasoning patterns than lexical overlap), or hybrid graph-text systems that exploit complementary signal.

\paragraph{Benchmark design.}
The train/test split ablation reveals that standard leave-one-out evaluation \emph{underestimates} real-world degradation by $\sim$5pp.
We recommend that future retrieval benchmarks adopt temporal splits as the default evaluation protocol.

\section{Dataset Release}
\label{sec:release}

UA-StatuteRetrieval is available at \url{https://huggingface.co/datasets/overthelex/ua-statute-retrieval}:

\begin{itemize}[nosep,leftmargin=*]
  \item \texttt{article\_retrieval\_performance.parquet}: per-article metrics for 3{,}667 articles.
  \item \texttt{temporal\_metrics.parquet}: year-by-year metrics (2007--2026), all methods.
  \item \texttt{difficulty\_stratification.csv}: per-bin breakdowns.
\end{itemize}

The R evaluation scripts are released alongside the paper.
The underlying 396M codex citation records remain proprietary; the co-citation graph is available separately~\citep{ovcharov2025citation}.

\section{Limitations}
\label{sec:limitations}

\begin{enumerate}[nosep,leftmargin=*]
  \item \emph{Link prediction, not retrieval}: our leave-one-out protocol measures citation prediction (given partial citations, recover the missing one), which is a proxy for statute retrieval but not the same task. Practitioners start from case facts, not from partial citation sets.
  \item \emph{Codex articles only}: specific laws (by number/date) are not covered.
  \item \emph{Citation $\neq$ relevance}: ground truth conflates procedural and substantive citations.
  \item \emph{Single jurisdiction}: results may not generalize to common-law systems where stare decisis creates different citation dynamics.
  \item \emph{No dense retrieval baseline}: our text baseline is BM25 (lexical); the embedding drift experiment uses E5-large for analysis but not as a retrieval method. Dense retrieval (E5, BGE-M3) may show different temporal dynamics than BM25.
  \item \emph{Early-year anomalies}: 2007 contains retrospective imports (15 cites/case vs.\ 4 average), and 2009 has only 52K decisions due to political crisis. Both outliers are retained but noted.
\end{enumerate}

\section{Conclusion}
\label{sec:conclusion}

We demonstrated that co-citation predictability in legal statute retrieval is \emph{non-stationary}: it decays 33--47\% over 12 years depending on evaluation protocol.
The decay is genuine (persists on fixed articles, strengthens without data leakage), non-uniform (concentrates in civil/administrative domains and mid-frequency articles), and practically significant ($\sim$4\% MRR/year during 2012--2018, slowing to $\sim$1\%/year after 2020).
BM25 text retrieval decays even faster (31\%), and embedding drift analysis reveals that the underlying cause is semantic evolution: the legal context in which articles are cited shifts 4.3\% over 12 years, with civil procedure drifting twice as fast as criminal law.

These findings challenge the assumption that legal retrieval signals -- whether structural or textual -- are temporally stable.
The released benchmark, with per-article difficulty scores, 20 annual snapshots, and embedding drift measurements, enables direct measurement of temporal robustness for any retrieval approach.

\bibliographystyle{plainnat}
\bibliography{references}

\end{document}

%% file: figures/fig1_difficulty_curve.tex
\begin{tikzpicture}[x=1pt,y=1pt]
\definecolor{fillColor}{RGB}{255,255,255}
\path[use as bounding box,fill=fillColor,fill opacity=0.00] (0,0) rectangle (397.48,231.26);
\begin{scope}
\path[clip] (  0.00,  0.00) rectangle (397.48,231.26);
\definecolor{fillColor}{RGB}{255,255,255}

\path[fill=fillColor] (  0.00,  0.00) rectangle (397.48,231.26);
\end{scope}
\begin{scope}
\path[clip] ( 31.05, 27.90) rectangle (392.48,226.26);
\definecolor{drawColor}{gray}{0.92}

\path[draw=drawColor,line width= 0.5pt,line join=round] ( 31.05, 36.91) --
	(392.48, 36.91);

\path[draw=drawColor,line width= 0.5pt,line join=round] ( 31.05, 72.98) --
	(392.48, 72.98);

\path[draw=drawColor,line width= 0.5pt,line join=round] ( 31.05,109.05) --
	(392.48,109.05);

\path[draw=drawColor,line width= 0.5pt,line join=round] ( 31.05,145.11) --
	(392.48,145.11);

\path[draw=drawColor,line width= 0.5pt,line join=round] ( 31.05,181.18) --
	(392.48,181.18);

\path[draw=drawColor,line width= 0.5pt,line join=round] ( 31.05,217.25) --
	(392.48,217.25);

\path[draw=drawColor,line width= 0.5pt,line join=round] (104.03, 27.90) --
	(104.03,226.26);

\path[draw=drawColor,line width= 0.5pt,line join=round] (191.09, 27.90) --
	(191.09,226.26);

\path[draw=drawColor,line width= 0.5pt,line join=round] (278.14, 27.90) --
	(278.14,226.26);

\path[draw=drawColor,line width= 0.5pt,line join=round] (365.20, 27.90) --
	(365.20,226.26);
\definecolor{fillColor}{RGB}{44,123,182}

\path[fill=fillColor] ( 47.48, 40.51) circle (  2.68);

\path[fill=fillColor] ( 64.73, 42.19) circle (  2.68);

\path[fill=fillColor] ( 81.94, 38.40) circle (  2.68);

\path[fill=fillColor] ( 97.42, 38.62) circle (  2.68);

\path[fill=fillColor] (115.09, 38.70) circle (  2.68);

\path[fill=fillColor] (132.49, 38.42) circle (  2.68);

\path[fill=fillColor] (150.09, 39.32) circle (  2.68);

\path[fill=fillColor] (167.29, 40.22) circle (  2.68);

\path[fill=fillColor] (184.20, 43.56) circle (  2.68);

\path[fill=fillColor] (202.39, 44.62) circle (  2.68);

\path[fill=fillColor] (218.80, 44.48) circle (  2.68);

\path[fill=fillColor] (236.98, 46.82) circle (  2.68);

\path[fill=fillColor] (254.10, 55.64) circle (  2.68);

\path[fill=fillColor] (271.45, 52.22) circle (  2.68);

\path[fill=fillColor] (288.39, 63.88) circle (  2.68);

\path[fill=fillColor] (305.63, 66.09) circle (  2.68);

\path[fill=fillColor] (322.61, 84.05) circle (  2.68);

\path[fill=fillColor] (340.94,109.49) circle (  2.68);

\path[fill=fillColor] (358.10, 97.95) circle (  2.68);

\path[fill=fillColor] (376.06,137.40) circle (  2.68);
\definecolor{fillColor}{RGB}{215,25,28}

\path[fill=fillColor] ( 47.48, 47.52) --
	( 51.09, 41.27) --
	( 43.87, 41.27) --
	cycle;

\path[fill=fillColor] ( 64.73, 47.17) --
	( 68.34, 40.92) --
	( 61.12, 40.92) --
	cycle;

\path[fill=fillColor] ( 81.94, 42.54) --
	( 85.55, 36.29) --
	( 78.33, 36.29) --
	cycle;

\path[fill=fillColor] ( 97.42, 42.65) --
	(101.03, 36.40) --
	( 93.81, 36.40) --
	cycle;

\path[fill=fillColor] (115.09, 42.63) --
	(118.70, 36.38) --
	(111.49, 36.38) --
	cycle;

\path[fill=fillColor] (132.49, 41.88) --
	(136.10, 35.63) --
	(128.88, 35.63) --
	cycle;

\path[fill=fillColor] (150.09, 43.42) --
	(153.70, 37.17) --
	(146.48, 37.17) --
	cycle;

\path[fill=fillColor] (167.29, 44.36) --
	(170.90, 38.11) --
	(163.68, 38.11) --
	cycle;

\path[fill=fillColor] (184.20, 47.15) --
	(187.81, 40.90) --
	(180.59, 40.90) --
	cycle;

\path[fill=fillColor] (202.39, 49.33) --
	(206.00, 43.08) --
	(198.78, 43.08) --
	cycle;

\path[fill=fillColor] (218.80, 49.83) --
	(222.41, 43.57) --
	(215.19, 43.57) --
	cycle;

\path[fill=fillColor] (236.98, 52.49) --
	(240.59, 46.24) --
	(233.37, 46.24) --
	cycle;

\path[fill=fillColor] (254.10, 65.09) --
	(257.71, 58.84) --
	(250.49, 58.84) --
	cycle;

\path[fill=fillColor] (271.45, 58.24) --
	(275.05, 51.99) --
	(267.84, 51.99) --
	cycle;

\path[fill=fillColor] (288.39, 80.50) --
	(292.00, 74.25) --
	(284.78, 74.25) --
	cycle;

\path[fill=fillColor] (305.63, 80.07) --
	(309.24, 73.82) --
	(302.02, 73.82) --
	cycle;

\path[fill=fillColor] (322.61,106.62) --
	(326.22,100.37) --
	(319.00,100.37) --
	cycle;

\path[fill=fillColor] (340.94,152.49) --
	(344.55,146.24) --
	(337.33,146.24) --
	cycle;

\path[fill=fillColor] (358.10,151.27) --
	(361.71,145.02) --
	(354.49,145.02) --
	cycle;

\path[fill=fillColor] (376.06,186.25) --
	(379.66,180.00) --
	(372.45,180.00) --
	cycle;
\definecolor{drawColor}{RGB}{44,123,182}

\path[draw=drawColor,line width= 0.8pt,line join=round] ( 47.48, 40.51) --
	( 64.73, 42.19) --
	( 81.94, 38.40) --
	( 97.42, 38.62) --
	(115.09, 38.70) --
	(132.49, 38.42) --
	(150.09, 39.32) --
	(167.29, 40.22) --
	(184.20, 43.56) --
	(202.39, 44.62) --
	(218.80, 44.48) --
	(236.98, 46.82) --
	(254.10, 55.64) --
	(271.45, 52.22) --
	(288.39, 63.88) --
	(305.63, 66.09) --
	(322.61, 84.05) --
	(340.94,109.49) --
	(358.10, 97.95) --
	(376.06,137.40);
\definecolor{drawColor}{RGB}{215,25,28}

\path[draw=drawColor,line width= 0.8pt,line join=round] ( 47.48, 43.35) --
	( 64.73, 43.00) --
	( 81.94, 38.38) --
	( 97.42, 38.48) --
	(115.09, 38.46) --
	(132.49, 37.71) --
	(150.09, 39.26) --
	(167.29, 40.19) --
	(184.20, 42.98) --
	(202.39, 45.17) --
	(218.80, 45.66) --
	(236.98, 48.32) --
	(254.10, 60.92) --
	(271.45, 54.07) --
	(288.39, 76.33) --
	(305.63, 75.90) --
	(322.61,102.45) --
	(340.94,148.32) --
	(358.10,147.10) --
	(376.06,182.08);
\end{scope}
\begin{scope}
\path[clip] (  0.00,  0.00) rectangle (397.48,231.26);
\definecolor{drawColor}{gray}{0.30}

\node[text=drawColor,anchor=base east,inner sep=0pt, outer sep=0pt, scale=  0.80] at ( 26.55, 34.16) {0.0};

\node[text=drawColor,anchor=base east,inner sep=0pt, outer sep=0pt, scale=  0.80] at ( 26.55, 70.22) {0.2};

\node[text=drawColor,anchor=base east,inner sep=0pt, outer sep=0pt, scale=  0.80] at ( 26.55,106.29) {0.4};

\node[text=drawColor,anchor=base east,inner sep=0pt, outer sep=0pt, scale=  0.80] at ( 26.55,142.36) {0.6};

\node[text=drawColor,anchor=base east,inner sep=0pt, outer sep=0pt, scale=  0.80] at ( 26.55,178.43) {0.8};

\node[text=drawColor,anchor=base east,inner sep=0pt, outer sep=0pt, scale=  0.80] at ( 26.55,214.49) {1.0};
\end{scope}
\begin{scope}
\path[clip] (  0.00,  0.00) rectangle (397.48,231.26);
\definecolor{drawColor}{gray}{0.30}

\node[text=drawColor,anchor=base,inner sep=0pt, outer sep=0pt, scale=  0.80] at (104.03, 17.89) {100};

\node[text=drawColor,anchor=base,inner sep=0pt, outer sep=0pt, scale=  0.80] at (191.09, 17.89) {1K};

\node[text=drawColor,anchor=base,inner sep=0pt, outer sep=0pt, scale=  0.80] at (278.14, 17.89) {10K};

\node[text=drawColor,anchor=base,inner sep=0pt, outer sep=0pt, scale=  0.80] at (365.20, 17.89) {100K};
\end{scope}
\begin{scope}
\path[clip] (  0.00,  0.00) rectangle (397.48,231.26);
\definecolor{drawColor}{RGB}{0,0,0}

\node[text=drawColor,anchor=base,inner sep=0pt, outer sep=0pt, scale=  1.00] at (211.77,  6.94) {Article citation frequency (log scale)};
\end{scope}
\begin{scope}
\path[clip] (  0.00,  0.00) rectangle (397.48,231.26);
\definecolor{drawColor}{RGB}{0,0,0}

\node[text=drawColor,rotate= 90.00,anchor=base,inner sep=0pt, outer sep=0pt, scale=  1.00] at ( 11.89,127.08) {Retrieval performance};
\end{scope}
\begin{scope}
\path[clip] (  0.00,  0.00) rectangle (397.48,231.26);
\definecolor{fillColor}{RGB}{255,255,255}

\path[fill=fillColor] ( 68.13,177.05) rectangle (174.69,215.96);
\end{scope}
\begin{scope}
\path[clip] (  0.00,  0.00) rectangle (397.48,231.26);
\definecolor{fillColor}{RGB}{44,123,182}

\path[fill=fillColor] ( 80.35,203.74) circle (  2.68);
\definecolor{drawColor}{RGB}{44,123,182}

\path[draw=drawColor,line width= 0.8pt,line join=round] ( 74.57,203.74) -- ( 86.13,203.74);
\end{scope}
\begin{scope}
\path[clip] (  0.00,  0.00) rectangle (397.48,231.26);
\definecolor{fillColor}{RGB}{215,25,28}

\path[fill=fillColor] ( 80.35,193.45) --
	( 83.96,187.20) --
	( 76.74,187.20) --
	cycle;
\definecolor{drawColor}{RGB}{215,25,28}

\path[draw=drawColor,line width= 0.8pt,line join=round] ( 74.57,189.28) -- ( 86.13,189.28);
\end{scope}
\begin{scope}
\path[clip] (  0.00,  0.00) rectangle (397.48,231.26);
\definecolor{drawColor}{RGB}{0,0,0}

\node[text=drawColor,anchor=base west,inner sep=0pt, outer sep=0pt, scale=  0.80] at ( 92.58,200.98) {MRR (Adamic-Adar)};
\end{scope}
\begin{scope}
\path[clip] (  0.00,  0.00) rectangle (397.48,231.26);
\definecolor{drawColor}{RGB}{0,0,0}

\node[text=drawColor,anchor=base west,inner sep=0pt, outer sep=0pt, scale=  0.80] at ( 92.58,186.53) {Hit@5 (Adamic-Adar)};
\end{scope}
\end{tikzpicture}

%% file: figures/fig3_temporal_curve.tex
\begin{tikzpicture}[x=1pt,y=1pt]
\definecolor{fillColor}{RGB}{255,255,255}
\path[use as bounding box,fill=fillColor,fill opacity=0.00] (0,0) rectangle (397.48,216.81);
\begin{scope}
\path[clip] (  0.00,  0.00) rectangle (397.48,216.81);
\definecolor{fillColor}{RGB}{255,255,255}

\path[fill=fillColor] ( -0.00,  0.00) rectangle (397.48,216.81);
\end{scope}
\begin{scope}
\path[clip] ( 31.05, 27.90) rectangle (392.48,211.81);
\definecolor{drawColor}{gray}{0.92}

\path[draw=drawColor,line width= 0.5pt,line join=round] ( 31.05, 36.26) --
	(392.48, 36.26);

\path[draw=drawColor,line width= 0.5pt,line join=round] ( 31.05, 64.12) --
	(392.48, 64.12);

\path[draw=drawColor,line width= 0.5pt,line join=round] ( 31.05, 91.99) --
	(392.48, 91.99);

\path[draw=drawColor,line width= 0.5pt,line join=round] ( 31.05,119.85) --
	(392.48,119.85);

\path[draw=drawColor,line width= 0.5pt,line join=round] ( 31.05,147.72) --
	(392.48,147.72);

\path[draw=drawColor,line width= 0.5pt,line join=round] ( 31.05,175.58) --
	(392.48,175.58);

\path[draw=drawColor,line width= 0.5pt,line join=round] ( 31.05,203.45) --
	(392.48,203.45);

\path[draw=drawColor,line width= 0.5pt,line join=round] ( 64.77, 27.90) --
	( 64.77,211.81);

\path[draw=drawColor,line width= 0.5pt,line join=round] ( 99.36, 27.90) --
	( 99.36,211.81);

\path[draw=drawColor,line width= 0.5pt,line join=round] (133.95, 27.90) --
	(133.95,211.81);

\path[draw=drawColor,line width= 0.5pt,line join=round] (168.53, 27.90) --
	(168.53,211.81);

\path[draw=drawColor,line width= 0.5pt,line join=round] (203.12, 27.90) --
	(203.12,211.81);

\path[draw=drawColor,line width= 0.5pt,line join=round] (237.71, 27.90) --
	(237.71,211.81);

\path[draw=drawColor,line width= 0.5pt,line join=round] (272.30, 27.90) --
	(272.30,211.81);

\path[draw=drawColor,line width= 0.5pt,line join=round] (306.88, 27.90) --
	(306.88,211.81);

\path[draw=drawColor,line width= 0.5pt,line join=round] (341.47, 27.90) --
	(341.47,211.81);

\path[draw=drawColor,line width= 0.5pt,line join=round] (376.06, 27.90) --
	(376.06,211.81);
\definecolor{drawColor}{RGB}{44,123,182}

\path[draw=drawColor,line width= 0.9pt,line join=round] ( 47.48,130.30) --
	( 64.77,153.31) --
	( 82.07,192.22) --
	( 99.36,135.38) --
	(116.65,154.47) --
	(133.95,165.80) --
	(151.24,143.76) --
	(168.53,138.79) --
	(185.83,132.93) --
	(203.12,134.59) --
	(220.41,129.81) --
	(237.71,122.48) --
	(255.00,122.76) --
	(272.30,117.56) --
	(289.59,114.99) --
	(306.88,114.24) --
	(324.18,115.41) --
	(341.47,112.06) --
	(358.76,112.27) --
	(376.06,114.89);
\definecolor{drawColor}{RGB}{253,174,97}

\path[draw=drawColor,line width= 0.9pt,line join=round] ( 47.48,129.31) --
	( 64.77,152.30) --
	( 82.07,190.19) --
	( 99.36,133.83) --
	(116.65,149.76) --
	(133.95,164.90) --
	(151.24,141.48) --
	(168.53,135.98) --
	(185.83,131.15) --
	(203.12,132.42) --
	(220.41,126.99) --
	(237.71,120.20) --
	(255.00,120.65) --
	(272.30,115.47) --
	(289.59,113.08) --
	(306.88,112.13) --
	(324.18,113.67) --
	(341.47,110.49) --
	(358.76,110.75) --
	(376.06,113.41);
\definecolor{drawColor}{RGB}{215,25,28}

\path[draw=drawColor,line width= 0.9pt,line join=round] ( 47.48, 88.70) --
	( 64.77, 86.70) --
	( 82.07,126.97) --
	( 99.36, 71.79) --
	(116.65, 86.51) --
	(133.95, 98.73) --
	(151.24, 59.98) --
	(168.53, 77.88) --
	(185.83, 61.99) --
	(203.12, 65.76) --
	(220.41, 61.43) --
	(237.71, 56.13) --
	(255.00, 54.01) --
	(272.30, 54.12) --
	(289.59, 53.11) --
	(306.88, 49.45) --
	(324.18, 49.12) --
	(341.47, 52.58) --
	(358.76, 54.50) --
	(376.06, 60.28);
\definecolor{fillColor}{RGB}{253,174,97}

\path[fill=fillColor] ( 47.48,132.81) --
	( 50.51,127.56) --
	( 44.45,127.56) --
	cycle;
\definecolor{fillColor}{RGB}{44,123,182}

\path[fill=fillColor] ( 47.48,130.30) circle (  2.25);
\definecolor{fillColor}{RGB}{215,25,28}

\path[fill=fillColor] ( 45.23, 86.45) --
	( 49.73, 86.45) --
	( 49.73, 90.95) --
	( 45.23, 90.95) --
	cycle;
\definecolor{fillColor}{RGB}{253,174,97}

\path[fill=fillColor] ( 64.77,155.80) --
	( 67.80,150.55) --
	( 61.74,150.55) --
	cycle;
\definecolor{fillColor}{RGB}{44,123,182}

\path[fill=fillColor] ( 64.77,153.31) circle (  2.25);
\definecolor{fillColor}{RGB}{215,25,28}

\path[fill=fillColor] ( 62.52, 84.45) --
	( 67.02, 84.45) --
	( 67.02, 88.95) --
	( 62.52, 88.95) --
	cycle;
\definecolor{fillColor}{RGB}{253,174,97}

\path[fill=fillColor] ( 82.07,193.69) --
	( 85.10,188.44) --
	( 79.04,188.44) --
	cycle;
\definecolor{fillColor}{RGB}{44,123,182}

\path[fill=fillColor] ( 82.07,192.22) circle (  2.25);
\definecolor{fillColor}{RGB}{215,25,28}

\path[fill=fillColor] ( 79.82,124.72) --
	( 84.32,124.72) --
	( 84.32,129.22) --
	( 79.82,129.22) --
	cycle;
\definecolor{fillColor}{RGB}{253,174,97}

\path[fill=fillColor] ( 99.36,137.33) --
	(102.39,132.08) --
	( 96.33,132.08) --
	cycle;
\definecolor{fillColor}{RGB}{44,123,182}

\path[fill=fillColor] ( 99.36,135.38) circle (  2.25);
\definecolor{fillColor}{RGB}{215,25,28}

\path[fill=fillColor] ( 97.11, 69.54) --
	(101.61, 69.54) --
	(101.61, 74.04) --
	( 97.11, 74.04) --
	cycle;
\definecolor{fillColor}{RGB}{253,174,97}

\path[fill=fillColor] (116.65,153.26) --
	(119.69,148.01) --
	(113.62,148.01) --
	cycle;
\definecolor{fillColor}{RGB}{44,123,182}

\path[fill=fillColor] (116.65,154.47) circle (  2.25);
\definecolor{fillColor}{RGB}{215,25,28}

\path[fill=fillColor] (114.40, 84.26) --
	(118.90, 84.26) --
	(118.90, 88.76) --
	(114.40, 88.76) --
	cycle;
\definecolor{fillColor}{RGB}{253,174,97}

\path[fill=fillColor] (133.95,168.40) --
	(136.98,163.15) --
	(130.92,163.15) --
	cycle;
\definecolor{fillColor}{RGB}{44,123,182}

\path[fill=fillColor] (133.95,165.80) circle (  2.25);
\definecolor{fillColor}{RGB}{215,25,28}

\path[fill=fillColor] (131.70, 96.48) --
	(136.20, 96.48) --
	(136.20,100.98) --
	(131.70,100.98) --
	cycle;
\definecolor{fillColor}{RGB}{253,174,97}

\path[fill=fillColor] (151.24,144.98) --
	(154.27,139.73) --
	(148.21,139.73) --
	cycle;
\definecolor{fillColor}{RGB}{44,123,182}

\path[fill=fillColor] (151.24,143.76) circle (  2.25);
\definecolor{fillColor}{RGB}{215,25,28}

\path[fill=fillColor] (148.99, 57.73) --
	(153.49, 57.73) --
	(153.49, 62.23) --
	(148.99, 62.23) --
	cycle;
\definecolor{fillColor}{RGB}{253,174,97}

\path[fill=fillColor] (168.53,139.48) --
	(171.57,134.23) --
	(165.50,134.23) --
	cycle;
\definecolor{fillColor}{RGB}{44,123,182}

\path[fill=fillColor] (168.53,138.79) circle (  2.25);
\definecolor{fillColor}{RGB}{215,25,28}

\path[fill=fillColor] (166.28, 75.63) --
	(170.79, 75.63) --
	(170.79, 80.13) --
	(166.28, 80.13) --
	cycle;
\definecolor{fillColor}{RGB}{253,174,97}

\path[fill=fillColor] (185.83,134.65) --
	(188.86,129.40) --
	(182.80,129.40) --
	cycle;
\definecolor{fillColor}{RGB}{44,123,182}

\path[fill=fillColor] (185.83,132.93) circle (  2.25);
\definecolor{fillColor}{RGB}{215,25,28}

\path[fill=fillColor] (183.58, 59.74) --
	(188.08, 59.74) --
	(188.08, 64.24) --
	(183.58, 64.24) --
	cycle;
\definecolor{fillColor}{RGB}{253,174,97}

\path[fill=fillColor] (203.12,135.92) --
	(206.15,130.67) --
	(200.09,130.67) --
	cycle;
\definecolor{fillColor}{RGB}{44,123,182}

\path[fill=fillColor] (203.12,134.59) circle (  2.25);
\definecolor{fillColor}{RGB}{215,25,28}

\path[fill=fillColor] (200.87, 63.51) --
	(205.37, 63.51) --
	(205.37, 68.01) --
	(200.87, 68.01) --
	cycle;
\definecolor{fillColor}{RGB}{253,174,97}

\path[fill=fillColor] (220.41,130.49) --
	(223.45,125.24) --
	(217.38,125.24) --
	cycle;
\definecolor{fillColor}{RGB}{44,123,182}

\path[fill=fillColor] (220.41,129.81) circle (  2.25);
\definecolor{fillColor}{RGB}{215,25,28}

\path[fill=fillColor] (218.16, 59.18) --
	(222.67, 59.18) --
	(222.67, 63.68) --
	(218.16, 63.68) --
	cycle;
\definecolor{fillColor}{RGB}{253,174,97}

\path[fill=fillColor] (237.71,123.70) --
	(240.74,118.45) --
	(234.68,118.45) --
	cycle;
\definecolor{fillColor}{RGB}{44,123,182}

\path[fill=fillColor] (237.71,122.48) circle (  2.25);
\definecolor{fillColor}{RGB}{215,25,28}

\path[fill=fillColor] (235.46, 53.88) --
	(239.96, 53.88) --
	(239.96, 58.38) --
	(235.46, 58.38) --
	cycle;
\definecolor{fillColor}{RGB}{253,174,97}

\path[fill=fillColor] (255.00,124.15) --
	(258.03,118.90) --
	(251.97,118.90) --
	cycle;
\definecolor{fillColor}{RGB}{44,123,182}

\path[fill=fillColor] (255.00,122.76) circle (  2.25);
\definecolor{fillColor}{RGB}{215,25,28}

\path[fill=fillColor] (252.75, 51.76) --
	(257.25, 51.76) --
	(257.25, 56.26) --
	(252.75, 56.26) --
	cycle;
\definecolor{fillColor}{RGB}{253,174,97}

\path[fill=fillColor] (272.30,118.97) --
	(275.33,113.72) --
	(269.26,113.72) --
	cycle;
\definecolor{fillColor}{RGB}{44,123,182}

\path[fill=fillColor] (272.30,117.56) circle (  2.25);
\definecolor{fillColor}{RGB}{215,25,28}

\path[fill=fillColor] (270.04, 51.87) --
	(274.55, 51.87) --
	(274.55, 56.37) --
	(270.04, 56.37) --
	cycle;
\definecolor{fillColor}{RGB}{253,174,97}

\path[fill=fillColor] (289.59,116.58) --
	(292.62,111.33) --
	(286.56,111.33) --
	cycle;
\definecolor{fillColor}{RGB}{44,123,182}

\path[fill=fillColor] (289.59,114.99) circle (  2.25);
\definecolor{fillColor}{RGB}{215,25,28}

\path[fill=fillColor] (287.34, 50.86) --
	(291.84, 50.86) --
	(291.84, 55.36) --
	(287.34, 55.36) --
	cycle;
\definecolor{fillColor}{RGB}{253,174,97}

\path[fill=fillColor] (306.88,115.63) --
	(309.91,110.38) --
	(303.85,110.38) --
	cycle;
\definecolor{fillColor}{RGB}{44,123,182}

\path[fill=fillColor] (306.88,114.24) circle (  2.25);
\definecolor{fillColor}{RGB}{215,25,28}

\path[fill=fillColor] (304.63, 47.20) --
	(309.13, 47.20) --
	(309.13, 51.70) --
	(304.63, 51.70) --
	cycle;
\definecolor{fillColor}{RGB}{253,174,97}

\path[fill=fillColor] (324.18,117.17) --
	(327.21,111.92) --
	(321.14,111.92) --
	cycle;
\definecolor{fillColor}{RGB}{44,123,182}

\path[fill=fillColor] (324.18,115.41) circle (  2.25);
\definecolor{fillColor}{RGB}{215,25,28}

\path[fill=fillColor] (321.92, 46.87) --
	(326.43, 46.87) --
	(326.43, 51.37) --
	(321.92, 51.37) --
	cycle;
\definecolor{fillColor}{RGB}{253,174,97}

\path[fill=fillColor] (341.47,113.99) --
	(344.50,108.74) --
	(338.44,108.74) --
	cycle;
\definecolor{fillColor}{RGB}{44,123,182}

\path[fill=fillColor] (341.47,112.06) circle (  2.25);
\definecolor{fillColor}{RGB}{215,25,28}

\path[fill=fillColor] (339.22, 50.33) --
	(343.72, 50.33) --
	(343.72, 54.84) --
	(339.22, 54.84) --
	cycle;
\definecolor{fillColor}{RGB}{253,174,97}

\path[fill=fillColor] (358.76,114.25) --
	(361.79,109.00) --
	(355.73,109.00) --
	cycle;
\definecolor{fillColor}{RGB}{44,123,182}

\path[fill=fillColor] (358.76,112.27) circle (  2.25);
\definecolor{fillColor}{RGB}{215,25,28}

\path[fill=fillColor] (356.51, 52.25) --
	(361.01, 52.25) --
	(361.01, 56.76) --
	(356.51, 56.76) --
	cycle;
\definecolor{fillColor}{RGB}{253,174,97}

\path[fill=fillColor] (376.06,116.91) --
	(379.09,111.66) --
	(373.02,111.66) --
	cycle;
\definecolor{fillColor}{RGB}{44,123,182}

\path[fill=fillColor] (376.06,114.89) circle (  2.25);
\definecolor{fillColor}{RGB}{215,25,28}

\path[fill=fillColor] (373.81, 58.03) --
	(378.31, 58.03) --
	(378.31, 62.53) --
	(373.81, 62.53) --
	cycle;
\definecolor{drawColor}{gray}{0.50}

\path[draw=drawColor,line width= 0.5pt,dash pattern=on 4pt off 4pt ,line join=round] (220.41, 27.90) -- (220.41,211.81);

\path[draw=drawColor,line width= 0.5pt,dash pattern=on 1pt off 3pt ,line join=round] (306.88, 27.90) -- (306.88,211.81);
\definecolor{drawColor}{gray}{0.40}

\node[text=drawColor,anchor=base west,inner sep=0pt, outer sep=0pt, scale=  0.71] at (225.60,178.71) {Reform};

\node[text=drawColor,anchor=base west,inner sep=0pt, outer sep=0pt, scale=  0.71] at (312.07,178.71) {Invasion};
\end{scope}
\begin{scope}
\path[clip] (  0.00,  0.00) rectangle (397.48,216.81);
\definecolor{drawColor}{gray}{0.30}

\node[text=drawColor,anchor=base east,inner sep=0pt, outer sep=0pt, scale=  0.80] at ( 26.55, 33.50) {0.0};

\node[text=drawColor,anchor=base east,inner sep=0pt, outer sep=0pt, scale=  0.80] at ( 26.55, 61.37) {0.1};

\node[text=drawColor,anchor=base east,inner sep=0pt, outer sep=0pt, scale=  0.80] at ( 26.55, 89.23) {0.2};

\node[text=drawColor,anchor=base east,inner sep=0pt, outer sep=0pt, scale=  0.80] at ( 26.55,117.10) {0.3};

\node[text=drawColor,anchor=base east,inner sep=0pt, outer sep=0pt, scale=  0.80] at ( 26.55,144.96) {0.4};

\node[text=drawColor,anchor=base east,inner sep=0pt, outer sep=0pt, scale=  0.80] at ( 26.55,172.83) {0.5};

\node[text=drawColor,anchor=base east,inner sep=0pt, outer sep=0pt, scale=  0.80] at ( 26.55,200.70) {0.6};
\end{scope}
\begin{scope}
\path[clip] (  0.00,  0.00) rectangle (397.48,216.81);
\definecolor{drawColor}{gray}{0.30}

\node[text=drawColor,anchor=base,inner sep=0pt, outer sep=0pt, scale=  0.80] at ( 64.77, 17.89) {2008};

\node[text=drawColor,anchor=base,inner sep=0pt, outer sep=0pt, scale=  0.80] at ( 99.36, 17.89) {2010};

\node[text=drawColor,anchor=base,inner sep=0pt, outer sep=0pt, scale=  0.80] at (133.95, 17.89) {2012};

\node[text=drawColor,anchor=base,inner sep=0pt, outer sep=0pt, scale=  0.80] at (168.53, 17.89) {2014};

\node[text=drawColor,anchor=base,inner sep=0pt, outer sep=0pt, scale=  0.80] at (203.12, 17.89) {2016};

\node[text=drawColor,anchor=base,inner sep=0pt, outer sep=0pt, scale=  0.80] at (237.71, 17.89) {2018};

\node[text=drawColor,anchor=base,inner sep=0pt, outer sep=0pt, scale=  0.80] at (272.30, 17.89) {2020};

\node[text=drawColor,anchor=base,inner sep=0pt, outer sep=0pt, scale=  0.80] at (306.88, 17.89) {2022};

\node[text=drawColor,anchor=base,inner sep=0pt, outer sep=0pt, scale=  0.80] at (341.47, 17.89) {2024};

\node[text=drawColor,anchor=base,inner sep=0pt, outer sep=0pt, scale=  0.80] at (376.06, 17.89) {2026};
\end{scope}
\begin{scope}
\path[clip] (  0.00,  0.00) rectangle (397.48,216.81);
\definecolor{drawColor}{RGB}{0,0,0}

\node[text=drawColor,anchor=base,inner sep=0pt, outer sep=0pt, scale=  1.00] at (211.77,  6.94) {Year};
\end{scope}
\begin{scope}
\path[clip] (  0.00,  0.00) rectangle (397.48,216.81);
\definecolor{drawColor}{RGB}{0,0,0}

\node[text=drawColor,rotate= 90.00,anchor=base,inner sep=0pt, outer sep=0pt, scale=  1.00] at ( 11.89,119.85) {MRR};
\end{scope}
\begin{scope}
\path[clip] (  0.00,  0.00) rectangle (397.48,216.81);
\definecolor{fillColor}{RGB}{255,255,255}

\path[fill=fillColor] (272.43,162.15) rectangle (367.97,206.29);
\end{scope}
\begin{scope}
\path[clip] (  0.00,  0.00) rectangle (397.48,216.81);
\definecolor{drawColor}{RGB}{44,123,182}

\path[draw=drawColor,line width= 0.9pt,line join=round] (278.57,195.60) -- (287.67,195.60);
\definecolor{fillColor}{RGB}{44,123,182}

\path[fill=fillColor] (283.12,195.60) circle (  2.25);
\end{scope}
\begin{scope}
\path[clip] (  0.00,  0.00) rectangle (397.48,216.81);
\definecolor{drawColor}{RGB}{253,174,97}

\path[draw=drawColor,line width= 0.9pt,line join=round] (278.57,184.22) -- (287.67,184.22);
\definecolor{fillColor}{RGB}{253,174,97}

\path[fill=fillColor] (283.12,187.72) --
	(286.15,182.47) --
	(280.09,182.47) --
	cycle;
\end{scope}
\begin{scope}
\path[clip] (  0.00,  0.00) rectangle (397.48,216.81);
\definecolor{drawColor}{RGB}{215,25,28}

\path[draw=drawColor,line width= 0.9pt,line join=round] (278.57,172.84) -- (287.67,172.84);
\definecolor{fillColor}{RGB}{215,25,28}

\path[fill=fillColor] (280.87,170.59) --
	(285.37,170.59) --
	(285.37,175.09) --
	(280.87,175.09) --
	cycle;
\end{scope}
\begin{scope}
\path[clip] (  0.00,  0.00) rectangle (397.48,216.81);
\definecolor{drawColor}{RGB}{0,0,0}

\node[text=drawColor,anchor=base west,inner sep=0pt, outer sep=0pt, scale=  0.80] at (293.81,192.85) {Adamic-Adar};
\end{scope}
\begin{scope}
\path[clip] (  0.00,  0.00) rectangle (397.48,216.81);
\definecolor{drawColor}{RGB}{0,0,0}

\node[text=drawColor,anchor=base west,inner sep=0pt, outer sep=0pt, scale=  0.80] at (293.81,181.47) {Common Neighbors};
\end{scope}
\begin{scope}
\path[clip] (  0.00,  0.00) rectangle (397.48,216.81);
\definecolor{drawColor}{RGB}{0,0,0}

\node[text=drawColor,anchor=base west,inner sep=0pt, outer sep=0pt, scale=  0.80] at (293.81,170.09) {Degree Baseline};
\end{scope}
\end{tikzpicture}

%% file: figures/fig4_year_difficulty_heatmap.tex
\begin{tikzpicture}[x=1pt,y=1pt]
\definecolor{fillColor}{RGB}{255,255,255}
\path[use as bounding box,fill=fillColor,fill opacity=0.00] (0,0) rectangle (397.48,216.81);
\begin{scope}
\path[clip] (  0.00,  0.00) rectangle (397.48,216.81);
\definecolor{fillColor}{RGB}{255,255,255}

\path[fill=fillColor] (  0.00,  0.00) rectangle (397.48,216.81);
\end{scope}
\begin{scope}
\path[clip] ( 71.49, 27.90) rectangle (342.81,211.81);
\definecolor{drawColor}{RGB}{255,255,255}
\definecolor{fillColor}{RGB}{130,163,187}

\path[draw=drawColor,line width= 0.3pt,fill=fillColor] ( 83.82,172.91) rectangle ( 96.15,208.27);
\definecolor{fillColor}{RGB}{254,215,156}

\path[draw=drawColor,line width= 0.3pt,fill=fillColor] ( 83.82, 66.80) rectangle ( 96.15,102.17);
\definecolor{fillColor}{RGB}{148,174,188}

\path[draw=drawColor,line width= 0.3pt,fill=fillColor] ( 83.82,137.54) rectangle ( 96.15,172.91);
\definecolor{fillColor}{RGB}{192,205,190}

\path[draw=drawColor,line width= 0.3pt,fill=fillColor] ( 83.82,102.17) rectangle ( 96.15,137.54);
\definecolor{fillColor}{RGB}{232,99,64}

\path[draw=drawColor,line width= 0.3pt,fill=fillColor] ( 83.82, 31.43) rectangle ( 96.15, 66.80);
\definecolor{fillColor}{RGB}{217,40,32}

\path[draw=drawColor,line width= 0.3pt,fill=fillColor] ( 96.15, 66.80) rectangle (108.48,102.17);
\definecolor{fillColor}{RGB}{255,228,167}

\path[draw=drawColor,line width= 0.3pt,fill=fillColor] ( 96.15,102.17) rectangle (108.48,137.54);
\definecolor{fillColor}{RGB}{183,199,189}

\path[draw=drawColor,line width= 0.3pt,fill=fillColor] ( 96.15,137.54) rectangle (108.48,172.91);
\definecolor{fillColor}{RGB}{153,178,188}

\path[draw=drawColor,line width= 0.3pt,fill=fillColor] ( 96.15,172.91) rectangle (108.48,208.27);
\definecolor{fillColor}{RGB}{236,240,191}

\path[draw=drawColor,line width= 0.3pt,fill=fillColor] (108.48,172.91) rectangle (120.82,208.27);
\definecolor{fillColor}{RGB}{223,67,44}

\path[draw=drawColor,line width= 0.3pt,fill=fillColor] (108.48,102.17) rectangle (120.82,137.54);
\definecolor{fillColor}{RGB}{225,231,191}

\path[draw=drawColor,line width= 0.3pt,fill=fillColor] (108.48,137.54) rectangle (120.82,172.91);
\definecolor{fillColor}{RGB}{222,61,41}

\path[draw=drawColor,line width= 0.3pt,fill=fillColor] (108.48, 66.80) rectangle (120.82,102.17);
\definecolor{fillColor}{RGB}{255,220,160}

\path[draw=drawColor,line width= 0.3pt,fill=fillColor] (120.82, 66.80) rectangle (133.15,102.17);
\definecolor{fillColor}{RGB}{182,198,189}

\path[draw=drawColor,line width= 0.3pt,fill=fillColor] (120.82,102.17) rectangle (133.15,137.54);
\definecolor{fillColor}{RGB}{222,62,42}

\path[draw=drawColor,line width= 0.3pt,fill=fillColor] (120.82, 31.43) rectangle (133.15, 66.80);
\definecolor{fillColor}{RGB}{147,174,188}

\path[draw=drawColor,line width= 0.3pt,fill=fillColor] (120.82,137.54) rectangle (133.15,172.91);
\definecolor{fillColor}{RGB}{138,167,187}

\path[draw=drawColor,line width= 0.3pt,fill=fillColor] (120.82,172.91) rectangle (133.15,208.27);
\definecolor{fillColor}{RGB}{246,157,107}

\path[draw=drawColor,line width= 0.3pt,fill=fillColor] (133.15, 66.80) rectangle (145.48,102.17);
\definecolor{fillColor}{RGB}{181,197,189}

\path[draw=drawColor,line width= 0.3pt,fill=fillColor] (133.15,172.91) rectangle (145.48,208.27);
\definecolor{fillColor}{RGB}{255,217,157}

\path[draw=drawColor,line width= 0.3pt,fill=fillColor] (133.15,102.17) rectangle (145.48,137.54);
\definecolor{fillColor}{RGB}{191,205,190}

\path[draw=drawColor,line width= 0.3pt,fill=fillColor] (133.15,137.54) rectangle (145.48,172.91);
\definecolor{fillColor}{RGB}{241,133,88}

\path[draw=drawColor,line width= 0.3pt,fill=fillColor] (145.48, 66.80) rectangle (157.82,102.17);
\definecolor{fillColor}{gray}{0.50}

\path[draw=drawColor,line width= 0.3pt,fill=fillColor] (145.48, 31.43) rectangle (157.82, 66.80);
\definecolor{fillColor}{RGB}{218,226,191}

\path[draw=drawColor,line width= 0.3pt,fill=fillColor] (145.48,102.17) rectangle (157.82,137.54);
\definecolor{fillColor}{RGB}{163,185,188}

\path[draw=drawColor,line width= 0.3pt,fill=fillColor] (145.48,137.54) rectangle (157.82,172.91);
\definecolor{fillColor}{RGB}{145,172,187}

\path[draw=drawColor,line width= 0.3pt,fill=fillColor] (145.48,172.91) rectangle (157.82,208.27);
\definecolor{fillColor}{RGB}{234,107,69}

\path[draw=drawColor,line width= 0.3pt,fill=fillColor] (157.82, 66.80) rectangle (170.15,102.17);
\definecolor{fillColor}{RGB}{249,250,191}

\path[draw=drawColor,line width= 0.3pt,fill=fillColor] (157.82,102.17) rectangle (170.15,137.54);
\definecolor{fillColor}{RGB}{135,166,187}

\path[draw=drawColor,line width= 0.3pt,fill=fillColor] (157.82,172.91) rectangle (170.15,208.27);
\definecolor{fillColor}{RGB}{156,179,188}

\path[draw=drawColor,line width= 0.3pt,fill=fillColor] (157.82,137.54) rectangle (170.15,172.91);
\definecolor{fillColor}{RGB}{228,83,53}

\path[draw=drawColor,line width= 0.3pt,fill=fillColor] (170.15, 31.43) rectangle (182.48, 66.80);
\definecolor{fillColor}{RGB}{255,217,157}

\path[draw=drawColor,line width= 0.3pt,fill=fillColor] (170.15, 66.80) rectangle (182.48,102.17);
\definecolor{fillColor}{RGB}{180,197,189}

\path[draw=drawColor,line width= 0.3pt,fill=fillColor] (170.15,102.17) rectangle (182.48,137.54);
\definecolor{fillColor}{RGB}{158,181,188}

\path[draw=drawColor,line width= 0.3pt,fill=fillColor] (170.15,137.54) rectangle (182.48,172.91);
\definecolor{fillColor}{RGB}{125,160,186}

\path[draw=drawColor,line width= 0.3pt,fill=fillColor] (170.15,172.91) rectangle (182.48,208.27);
\definecolor{fillColor}{RGB}{129,162,187}

\path[draw=drawColor,line width= 0.3pt,fill=fillColor] (182.48,172.91) rectangle (194.82,208.27);
\definecolor{fillColor}{RGB}{255,226,165}

\path[draw=drawColor,line width= 0.3pt,fill=fillColor] (182.48, 66.80) rectangle (194.82,102.17);
\definecolor{fillColor}{RGB}{221,55,38}

\path[draw=drawColor,line width= 0.3pt,fill=fillColor] (182.48, 31.43) rectangle (194.82, 66.80);
\definecolor{fillColor}{RGB}{192,206,190}

\path[draw=drawColor,line width= 0.3pt,fill=fillColor] (182.48,102.17) rectangle (194.82,137.54);
\definecolor{fillColor}{RGB}{146,173,188}

\path[draw=drawColor,line width= 0.3pt,fill=fillColor] (182.48,137.54) rectangle (194.82,172.91);
\definecolor{fillColor}{RGB}{238,120,78}

\path[draw=drawColor,line width= 0.3pt,fill=fillColor] (194.82, 31.43) rectangle (207.15, 66.80);
\definecolor{fillColor}{RGB}{240,243,191}

\path[draw=drawColor,line width= 0.3pt,fill=fillColor] (194.82, 66.80) rectangle (207.15,102.17);
\definecolor{fillColor}{RGB}{168,188,189}

\path[draw=drawColor,line width= 0.3pt,fill=fillColor] (194.82,102.17) rectangle (207.15,137.54);
\definecolor{fillColor}{RGB}{133,164,187}

\path[draw=drawColor,line width= 0.3pt,fill=fillColor] (194.82,137.54) rectangle (207.15,172.91);
\definecolor{fillColor}{RGB}{117,155,186}

\path[draw=drawColor,line width= 0.3pt,fill=fillColor] (194.82,172.91) rectangle (207.15,208.27);
\definecolor{fillColor}{RGB}{221,57,39}

\path[draw=drawColor,line width= 0.3pt,fill=fillColor] (207.15, 31.43) rectangle (219.48, 66.80);
\definecolor{fillColor}{RGB}{186,201,189}

\path[draw=drawColor,line width= 0.3pt,fill=fillColor] (207.15,102.17) rectangle (219.48,137.54);
\definecolor{fillColor}{RGB}{255,231,169}

\path[draw=drawColor,line width= 0.3pt,fill=fillColor] (207.15, 66.80) rectangle (219.48,102.17);
\definecolor{fillColor}{RGB}{134,165,187}

\path[draw=drawColor,line width= 0.3pt,fill=fillColor] (207.15,172.91) rectangle (219.48,208.27);
\definecolor{fillColor}{RGB}{139,168,187}

\path[draw=drawColor,line width= 0.3pt,fill=fillColor] (207.15,137.54) rectangle (219.48,172.91);
\definecolor{fillColor}{RGB}{255,238,176}

\path[draw=drawColor,line width= 0.3pt,fill=fillColor] (219.48, 66.80) rectangle (231.81,102.17);
\definecolor{fillColor}{RGB}{175,193,189}

\path[draw=drawColor,line width= 0.3pt,fill=fillColor] (219.48,102.17) rectangle (231.81,137.54);
\definecolor{fillColor}{RGB}{236,114,74}

\path[draw=drawColor,line width= 0.3pt,fill=fillColor] (219.48, 31.43) rectangle (231.81, 66.80);
\definecolor{fillColor}{RGB}{133,164,187}

\path[draw=drawColor,line width= 0.3pt,fill=fillColor] (219.48,137.54) rectangle (231.81,172.91);
\definecolor{fillColor}{RGB}{124,159,186}

\path[draw=drawColor,line width= 0.3pt,fill=fillColor] (219.48,172.91) rectangle (231.81,208.27);
\definecolor{fillColor}{RGB}{254,213,154}

\path[draw=drawColor,line width= 0.3pt,fill=fillColor] (231.81, 66.80) rectangle (244.15,102.17);
\definecolor{fillColor}{gray}{0.50}

\path[draw=drawColor,line width= 0.3pt,fill=fillColor] (231.81, 31.43) rectangle (244.15, 66.80);
\definecolor{fillColor}{RGB}{140,169,187}

\path[draw=drawColor,line width= 0.3pt,fill=fillColor] (231.81,137.54) rectangle (244.15,172.91);
\definecolor{fillColor}{RGB}{175,193,189}

\path[draw=drawColor,line width= 0.3pt,fill=fillColor] (231.81,102.17) rectangle (244.15,137.54);
\definecolor{fillColor}{RGB}{124,159,186}

\path[draw=drawColor,line width= 0.3pt,fill=fillColor] (231.81,172.91) rectangle (244.15,208.27);
\definecolor{fillColor}{RGB}{138,168,187}

\path[draw=drawColor,line width= 0.3pt,fill=fillColor] (244.15,137.54) rectangle (256.48,172.91);
\definecolor{fillColor}{RGB}{255,229,168}

\path[draw=drawColor,line width= 0.3pt,fill=fillColor] (244.15, 66.80) rectangle (256.48,102.17);
\definecolor{fillColor}{RGB}{227,80,52}

\path[draw=drawColor,line width= 0.3pt,fill=fillColor] (244.15, 31.43) rectangle (256.48, 66.80);
\definecolor{fillColor}{RGB}{177,195,189}

\path[draw=drawColor,line width= 0.3pt,fill=fillColor] (244.15,102.17) rectangle (256.48,137.54);
\definecolor{fillColor}{RGB}{131,163,187}

\path[draw=drawColor,line width= 0.3pt,fill=fillColor] (244.15,172.91) rectangle (256.48,208.27);
\definecolor{fillColor}{RGB}{255,231,169}

\path[draw=drawColor,line width= 0.3pt,fill=fillColor] (256.48, 66.80) rectangle (268.81,102.17);
\definecolor{fillColor}{RGB}{174,192,189}

\path[draw=drawColor,line width= 0.3pt,fill=fillColor] (256.48,102.17) rectangle (268.81,137.54);
\definecolor{fillColor}{RGB}{145,172,187}

\path[draw=drawColor,line width= 0.3pt,fill=fillColor] (256.48,137.54) rectangle (268.81,172.91);
\definecolor{fillColor}{gray}{0.50}

\path[draw=drawColor,line width= 0.3pt,fill=fillColor] (256.48, 31.43) rectangle (268.81, 66.80);
\definecolor{fillColor}{RGB}{121,157,186}

\path[draw=drawColor,line width= 0.3pt,fill=fillColor] (256.48,172.91) rectangle (268.81,208.27);
\definecolor{fillColor}{RGB}{255,219,159}

\path[draw=drawColor,line width= 0.3pt,fill=fillColor] (268.81, 66.80) rectangle (281.15,102.17);
\definecolor{fillColor}{RGB}{170,190,189}

\path[draw=drawColor,line width= 0.3pt,fill=fillColor] (268.81,102.17) rectangle (281.15,137.54);
\definecolor{fillColor}{RGB}{134,165,187}

\path[draw=drawColor,line width= 0.3pt,fill=fillColor] (268.81,137.54) rectangle (281.15,172.91);
\definecolor{fillColor}{RGB}{122,158,186}

\path[draw=drawColor,line width= 0.3pt,fill=fillColor] (268.81,172.91) rectangle (281.15,208.27);
\definecolor{fillColor}{RGB}{255,248,185}

\path[draw=drawColor,line width= 0.3pt,fill=fillColor] (281.15, 66.80) rectangle (293.48,102.17);
\definecolor{fillColor}{RGB}{234,106,68}

\path[draw=drawColor,line width= 0.3pt,fill=fillColor] (281.15, 31.43) rectangle (293.48, 66.80);
\definecolor{fillColor}{RGB}{158,181,188}

\path[draw=drawColor,line width= 0.3pt,fill=fillColor] (281.15,102.17) rectangle (293.48,137.54);
\definecolor{fillColor}{RGB}{121,157,186}

\path[draw=drawColor,line width= 0.3pt,fill=fillColor] (281.15,172.91) rectangle (293.48,208.27);
\definecolor{fillColor}{RGB}{129,162,187}

\path[draw=drawColor,line width= 0.3pt,fill=fillColor] (281.15,137.54) rectangle (293.48,172.91);
\definecolor{fillColor}{RGB}{244,246,191}

\path[draw=drawColor,line width= 0.3pt,fill=fillColor] (293.48, 66.80) rectangle (305.81,102.17);
\definecolor{fillColor}{RGB}{242,137,91}

\path[draw=drawColor,line width= 0.3pt,fill=fillColor] (293.48, 31.43) rectangle (305.81, 66.80);
\definecolor{fillColor}{RGB}{121,157,186}

\path[draw=drawColor,line width= 0.3pt,fill=fillColor] (293.48,172.91) rectangle (305.81,208.27);
\definecolor{fillColor}{RGB}{153,178,188}

\path[draw=drawColor,line width= 0.3pt,fill=fillColor] (293.48,102.17) rectangle (305.81,137.54);
\definecolor{fillColor}{RGB}{126,160,186}

\path[draw=drawColor,line width= 0.3pt,fill=fillColor] (293.48,137.54) rectangle (305.81,172.91);
\definecolor{fillColor}{RGB}{119,156,186}

\path[draw=drawColor,line width= 0.3pt,fill=fillColor] (305.81,172.91) rectangle (318.15,208.27);
\definecolor{fillColor}{RGB}{244,246,191}

\path[draw=drawColor,line width= 0.3pt,fill=fillColor] (305.81, 66.80) rectangle (318.15,102.17);
\definecolor{fillColor}{RGB}{242,139,93}

\path[draw=drawColor,line width= 0.3pt,fill=fillColor] (305.81, 31.43) rectangle (318.15, 66.80);
\definecolor{fillColor}{RGB}{153,178,188}

\path[draw=drawColor,line width= 0.3pt,fill=fillColor] (305.81,102.17) rectangle (318.15,137.54);
\definecolor{fillColor}{RGB}{128,161,186}

\path[draw=drawColor,line width= 0.3pt,fill=fillColor] (305.81,137.54) rectangle (318.15,172.91);
\definecolor{fillColor}{RGB}{243,245,191}

\path[draw=drawColor,line width= 0.3pt,fill=fillColor] (318.15, 66.80) rectangle (330.48,102.17);
\definecolor{fillColor}{RGB}{246,154,104}

\path[draw=drawColor,line width= 0.3pt,fill=fillColor] (318.15, 31.43) rectangle (330.48, 66.80);
\definecolor{fillColor}{RGB}{157,180,188}

\path[draw=drawColor,line width= 0.3pt,fill=fillColor] (318.15,102.17) rectangle (330.48,137.54);
\definecolor{fillColor}{RGB}{127,161,186}

\path[draw=drawColor,line width= 0.3pt,fill=fillColor] (318.15,137.54) rectangle (330.48,172.91);
\definecolor{fillColor}{RGB}{120,157,186}

\path[draw=drawColor,line width= 0.3pt,fill=fillColor] (318.15,172.91) rectangle (330.48,208.27);
\end{scope}
\begin{scope}
\path[clip] (  0.00,  0.00) rectangle (397.48,216.81);
\definecolor{drawColor}{gray}{0.30}

\node[text=drawColor,anchor=base east,inner sep=0pt, outer sep=0pt, scale=  0.80] at ( 66.99, 46.36) {Hub ($>$100K)};

\node[text=drawColor,anchor=base east,inner sep=0pt, outer sep=0pt, scale=  0.80] at ( 66.99, 81.73) {High (10K--100K)};

\node[text=drawColor,anchor=base east,inner sep=0pt, outer sep=0pt, scale=  0.80] at ( 66.99,117.10) {Mid (1K--10K)};

\node[text=drawColor,anchor=base east,inner sep=0pt, outer sep=0pt, scale=  0.80] at ( 66.99,152.47) {Low (100--1K)};

\node[text=drawColor,anchor=base east,inner sep=0pt, outer sep=0pt, scale=  0.80] at ( 66.99,187.83) {Rare ($<$100)};
\end{scope}
\begin{scope}
\path[clip] (  0.00,  0.00) rectangle (397.48,216.81);
\definecolor{drawColor}{gray}{0.30}

\node[text=drawColor,anchor=base,inner sep=0pt, outer sep=0pt, scale=  0.80] at (102.32, 17.89) {2008};

\node[text=drawColor,anchor=base,inner sep=0pt, outer sep=0pt, scale=  0.80] at (126.98, 17.89) {2010};

\node[text=drawColor,anchor=base,inner sep=0pt, outer sep=0pt, scale=  0.80] at (151.65, 17.89) {2012};

\node[text=drawColor,anchor=base,inner sep=0pt, outer sep=0pt, scale=  0.80] at (176.32, 17.89) {2014};

\node[text=drawColor,anchor=base,inner sep=0pt, outer sep=0pt, scale=  0.80] at (200.98, 17.89) {2016};

\node[text=drawColor,anchor=base,inner sep=0pt, outer sep=0pt, scale=  0.80] at (225.65, 17.89) {2018};

\node[text=drawColor,anchor=base,inner sep=0pt, outer sep=0pt, scale=  0.80] at (250.31, 17.89) {2020};

\node[text=drawColor,anchor=base,inner sep=0pt, outer sep=0pt, scale=  0.80] at (274.98, 17.89) {2022};

\node[text=drawColor,anchor=base,inner sep=0pt, outer sep=0pt, scale=  0.80] at (299.65, 17.89) {2024};

\node[text=drawColor,anchor=base,inner sep=0pt, outer sep=0pt, scale=  0.80] at (324.31, 17.89) {2026};
\end{scope}
\begin{scope}
\path[clip] (  0.00,  0.00) rectangle (397.48,216.81);
\definecolor{drawColor}{RGB}{0,0,0}

\node[text=drawColor,anchor=base,inner sep=0pt, outer sep=0pt, scale=  1.00] at (207.15,  6.94) {Year};
\end{scope}
\begin{scope}
\path[clip] (  0.00,  0.00) rectangle (397.48,216.81);
\definecolor{drawColor}{RGB}{0,0,0}

\node[text=drawColor,anchor=base west,inner sep=0pt, outer sep=0pt, scale=  1.00] at (357.81,175.81) {MRR};
\end{scope}
\begin{scope}
\path[clip] (  0.00,  0.00) rectangle (397.48,216.81);
\node[inner sep=0pt,outer sep=0pt,anchor=south west,rotate=  0.00] at (357.81,  56.03) {
	\pgfimage[width= 14.45pt,height=113.81pt,interpolate=true]{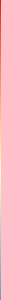}};
\end{scope}
\begin{scope}
\path[clip] (  0.00,  0.00) rectangle (397.48,216.81);
\definecolor{drawColor}{RGB}{255,255,255}

\path[draw=drawColor,line width= 0.2pt,line join=round] (369.37, 56.22) --
	(372.27, 56.22);

\path[draw=drawColor,line width= 0.2pt,line join=round] (369.37, 88.63) --
	(372.27, 88.63);

\path[draw=drawColor,line width= 0.2pt,line join=round] (369.37,121.04) --
	(372.27,121.04);

\path[draw=drawColor,line width= 0.2pt,line join=round] (369.37,153.45) --
	(372.27,153.45);

\path[draw=drawColor,line width= 0.2pt,line join=round] (360.70, 56.22) --
	(357.81, 56.22);

\path[draw=drawColor,line width= 0.2pt,line join=round] (360.70, 88.63) --
	(357.81, 88.63);

\path[draw=drawColor,line width= 0.2pt,line join=round] (360.70,121.04) --
	(357.81,121.04);

\path[draw=drawColor,line width= 0.2pt,line join=round] (360.70,153.45) --
	(357.81,153.45);
\end{scope}
\begin{scope}
\path[clip] (  0.00,  0.00) rectangle (397.48,216.81);
\definecolor{drawColor}{RGB}{0,0,0}

\node[text=drawColor,anchor=base west,inner sep=0pt, outer sep=0pt, scale=  0.80] at (377.27, 53.47) {0.0};

\node[text=drawColor,anchor=base west,inner sep=0pt, outer sep=0pt, scale=  0.80] at (377.27, 85.88) {0.2};

\node[text=drawColor,anchor=base west,inner sep=0pt, outer sep=0pt, scale=  0.80] at (377.27,118.28) {0.4};

\node[text=drawColor,anchor=base west,inner sep=0pt, outer sep=0pt, scale=  0.80] at (377.27,150.69) {0.6};
\end{scope}
\end{tikzpicture}

%% file: figures/fig6_per_codex_temporal.tex
\begin{tikzpicture}[x=1pt,y=1pt]
\definecolor{fillColor}{RGB}{255,255,255}
\path[use as bounding box,fill=fillColor,fill opacity=0.00] (0,0) rectangle (397.48,252.94);
\begin{scope}
\path[clip] (  0.00,  0.00) rectangle (397.48,252.94);
\definecolor{fillColor}{RGB}{255,255,255}

\path[fill=fillColor] (  0.00,  0.00) rectangle (397.48,252.94);
\end{scope}
\begin{scope}
\path[clip] ( 31.05, 75.66) rectangle (392.48,247.95);
\definecolor{drawColor}{gray}{0.92}

\path[draw=drawColor,line width= 0.5pt,line join=round] ( 31.05, 83.49) --
	(392.48, 83.49);

\path[draw=drawColor,line width= 0.5pt,line join=round] ( 31.05,105.86) --
	(392.48,105.86);

\path[draw=drawColor,line width= 0.5pt,line join=round] ( 31.05,128.24) --
	(392.48,128.24);

\path[draw=drawColor,line width= 0.5pt,line join=round] ( 31.05,150.61) --
	(392.48,150.61);

\path[draw=drawColor,line width= 0.5pt,line join=round] ( 31.05,172.99) --
	(392.48,172.99);

\path[draw=drawColor,line width= 0.5pt,line join=round] ( 31.05,195.36) --
	(392.48,195.36);

\path[draw=drawColor,line width= 0.5pt,line join=round] ( 31.05,217.74) --
	(392.48,217.74);

\path[draw=drawColor,line width= 0.5pt,line join=round] ( 31.05,240.11) --
	(392.48,240.11);

\path[draw=drawColor,line width= 0.5pt,line join=round] ( 64.77, 75.66) --
	( 64.77,247.95);

\path[draw=drawColor,line width= 0.5pt,line join=round] ( 99.36, 75.66) --
	( 99.36,247.95);

\path[draw=drawColor,line width= 0.5pt,line join=round] (133.95, 75.66) --
	(133.95,247.95);

\path[draw=drawColor,line width= 0.5pt,line join=round] (168.53, 75.66) --
	(168.53,247.95);

\path[draw=drawColor,line width= 0.5pt,line join=round] (203.12, 75.66) --
	(203.12,247.95);

\path[draw=drawColor,line width= 0.5pt,line join=round] (237.71, 75.66) --
	(237.71,247.95);

\path[draw=drawColor,line width= 0.5pt,line join=round] (272.30, 75.66) --
	(272.30,247.95);

\path[draw=drawColor,line width= 0.5pt,line join=round] (306.88, 75.66) --
	(306.88,247.95);

\path[draw=drawColor,line width= 0.5pt,line join=round] (341.47, 75.66) --
	(341.47,247.95);

\path[draw=drawColor,line width= 0.5pt,line join=round] (376.06, 75.66) --
	(376.06,247.95);
\definecolor{drawColor}{RGB}{228,26,28}

\path[draw=drawColor,line width= 0.9pt,line join=round] ( 47.48,178.93) --
	( 64.77,194.72) --
	( 82.07,214.30) --
	( 99.36,172.27) --
	(116.65,176.34) --
	(133.95,206.91) --
	(151.24,185.24) --
	(168.53,198.44) --
	(185.83,174.08) --
	(203.12,170.47) --
	(220.41,164.87) --
	(237.71,150.96) --
	(255.00,153.79) --
	(272.30,150.55) --
	(289.59,148.42) --
	(306.88,154.72) --
	(324.18,152.43) --
	(341.47,153.12) --
	(358.76,153.57) --
	(376.06,159.58);
\definecolor{drawColor}{RGB}{55,126,184}

\path[draw=drawColor,line width= 0.9pt,line join=round] ( 47.48, 99.46) --
	( 64.77,103.83) --
	( 82.07,152.66) --
	( 99.36, 99.24) --
	(116.65,106.92) --
	(133.95,105.49) --
	(151.24,106.66) --
	(168.53,107.64) --
	(185.83,107.06) --
	(203.12,102.55) --
	(220.41,101.76) --
	(237.71, 97.57) --
	(255.00, 97.88) --
	(272.30,100.46) --
	(289.59,102.21) --
	(306.88, 98.29) --
	(324.18,102.38) --
	(341.47,103.71) --
	(358.76,105.28) --
	(376.06,114.24);
\definecolor{drawColor}{RGB}{77,175,74}

\path[draw=drawColor,line width= 0.9pt,line join=round] ( 47.48,170.53) --
	( 64.77,175.48) --
	( 82.07,214.63) --
	( 99.36,177.27) --
	(116.65,201.93) --
	(133.95,190.60) --
	(151.24,184.98) --
	(168.53,179.04) --
	(185.83,181.59) --
	(203.12,179.25) --
	(220.41,174.02) --
	(237.71,156.51) --
	(255.00,157.76) --
	(272.30,157.32) --
	(289.59,155.83) --
	(306.88,152.84) --
	(324.18,154.70) --
	(341.47,151.28) --
	(358.76,151.42) --
	(376.06,153.31);
\definecolor{drawColor}{RGB}{152,78,163}

\path[draw=drawColor,line width= 0.9pt,line join=round] ( 47.48,136.83) --
	( 64.77,145.14) --
	( 82.07,141.03) --
	( 99.36,143.83) --
	(116.65,158.86) --
	(133.95,145.91) --
	(151.24,121.39) --
	(168.53,113.94) --
	(185.83,127.65) --
	(203.12,125.21) --
	(220.41,119.28) --
	(237.71,133.85) --
	(255.00,134.65) --
	(272.30,112.55) --
	(289.59,112.71) --
	(306.88,113.41) --
	(324.18,110.75) --
	(341.47,113.52) --
	(358.76,107.09) --
	(376.06,105.80);
\definecolor{drawColor}{RGB}{255,127,0}

\path[draw=drawColor,line width= 0.9pt,line join=round] ( 47.48,180.11) --
	( 64.77,199.56) --
	( 82.07,207.82) --
	( 99.36,198.59) --
	(116.65,219.18) --
	(133.95,200.24) --
	(151.24,183.40) --
	(168.53,196.10) --
	(185.83,173.85) --
	(203.12,184.99) --
	(220.41,181.39) --
	(237.71,185.93) --
	(255.00,183.13) --
	(272.30,174.18) --
	(289.59,171.58) --
	(306.88,165.95) --
	(324.18,167.36) --
	(341.47,166.58) --
	(358.76,167.64) --
	(376.06,166.30);
\definecolor{drawColor}{RGB}{228,26,28}
\definecolor{fillColor}{RGB}{228,26,28}

\path[draw=drawColor,line width= 0.3pt,line join=round,line cap=round,fill=fillColor] ( 47.48,178.93) circle (  1.61);
\definecolor{drawColor}{RGB}{255,127,0}
\definecolor{fillColor}{RGB}{255,127,0}

\path[draw=drawColor,line width= 0.3pt,line join=round,line cap=round,fill=fillColor] ( 47.48,180.11) circle (  1.61);
\definecolor{drawColor}{RGB}{77,175,74}
\definecolor{fillColor}{RGB}{77,175,74}

\path[draw=drawColor,line width= 0.3pt,line join=round,line cap=round,fill=fillColor] ( 47.48,170.53) circle (  1.61);
\definecolor{drawColor}{RGB}{152,78,163}
\definecolor{fillColor}{RGB}{152,78,163}

\path[draw=drawColor,line width= 0.3pt,line join=round,line cap=round,fill=fillColor] ( 47.48,136.83) circle (  1.61);
\definecolor{drawColor}{RGB}{55,126,184}
\definecolor{fillColor}{RGB}{55,126,184}

\path[draw=drawColor,line width= 0.3pt,line join=round,line cap=round,fill=fillColor] ( 47.48, 99.46) circle (  1.61);
\definecolor{drawColor}{RGB}{228,26,28}
\definecolor{fillColor}{RGB}{228,26,28}

\path[draw=drawColor,line width= 0.3pt,line join=round,line cap=round,fill=fillColor] ( 64.77,194.72) circle (  1.61);
\definecolor{drawColor}{RGB}{255,127,0}
\definecolor{fillColor}{RGB}{255,127,0}

\path[draw=drawColor,line width= 0.3pt,line join=round,line cap=round,fill=fillColor] ( 64.77,199.56) circle (  1.61);
\definecolor{drawColor}{RGB}{152,78,163}
\definecolor{fillColor}{RGB}{152,78,163}

\path[draw=drawColor,line width= 0.3pt,line join=round,line cap=round,fill=fillColor] ( 64.77,145.14) circle (  1.61);
\definecolor{drawColor}{RGB}{77,175,74}
\definecolor{fillColor}{RGB}{77,175,74}

\path[draw=drawColor,line width= 0.3pt,line join=round,line cap=round,fill=fillColor] ( 64.77,175.48) circle (  1.61);
\definecolor{drawColor}{RGB}{55,126,184}
\definecolor{fillColor}{RGB}{55,126,184}

\path[draw=drawColor,line width= 0.3pt,line join=round,line cap=round,fill=fillColor] ( 64.77,103.83) circle (  1.61);

\path[draw=drawColor,line width= 0.3pt,line join=round,line cap=round,fill=fillColor] ( 82.07,152.66) circle (  1.61);
\definecolor{drawColor}{RGB}{77,175,74}
\definecolor{fillColor}{RGB}{77,175,74}

\path[draw=drawColor,line width= 0.3pt,line join=round,line cap=round,fill=fillColor] ( 82.07,214.63) circle (  1.61);
\definecolor{drawColor}{RGB}{255,127,0}
\definecolor{fillColor}{RGB}{255,127,0}

\path[draw=drawColor,line width= 0.3pt,line join=round,line cap=round,fill=fillColor] ( 82.07,207.82) circle (  1.61);
\definecolor{drawColor}{RGB}{228,26,28}
\definecolor{fillColor}{RGB}{228,26,28}

\path[draw=drawColor,line width= 0.3pt,line join=round,line cap=round,fill=fillColor] ( 82.07,214.30) circle (  1.61);
\definecolor{drawColor}{RGB}{152,78,163}
\definecolor{fillColor}{RGB}{152,78,163}

\path[draw=drawColor,line width= 0.3pt,line join=round,line cap=round,fill=fillColor] ( 82.07,141.03) circle (  1.61);
\definecolor{drawColor}{RGB}{77,175,74}
\definecolor{fillColor}{RGB}{77,175,74}

\path[draw=drawColor,line width= 0.3pt,line join=round,line cap=round,fill=fillColor] ( 99.36,177.27) circle (  1.61);
\definecolor{drawColor}{RGB}{255,127,0}
\definecolor{fillColor}{RGB}{255,127,0}

\path[draw=drawColor,line width= 0.3pt,line join=round,line cap=round,fill=fillColor] ( 99.36,198.59) circle (  1.61);
\definecolor{drawColor}{RGB}{228,26,28}
\definecolor{fillColor}{RGB}{228,26,28}

\path[draw=drawColor,line width= 0.3pt,line join=round,line cap=round,fill=fillColor] ( 99.36,172.27) circle (  1.61);
\definecolor{drawColor}{RGB}{152,78,163}
\definecolor{fillColor}{RGB}{152,78,163}

\path[draw=drawColor,line width= 0.3pt,line join=round,line cap=round,fill=fillColor] ( 99.36,143.83) circle (  1.61);
\definecolor{drawColor}{RGB}{55,126,184}
\definecolor{fillColor}{RGB}{55,126,184}

\path[draw=drawColor,line width= 0.3pt,line join=round,line cap=round,fill=fillColor] ( 99.36, 99.24) circle (  1.61);
\definecolor{drawColor}{RGB}{77,175,74}
\definecolor{fillColor}{RGB}{77,175,74}

\path[draw=drawColor,line width= 0.3pt,line join=round,line cap=round,fill=fillColor] (116.65,201.93) circle (  1.61);
\definecolor{drawColor}{RGB}{255,127,0}
\definecolor{fillColor}{RGB}{255,127,0}

\path[draw=drawColor,line width= 0.3pt,line join=round,line cap=round,fill=fillColor] (116.65,219.18) circle (  1.61);
\definecolor{drawColor}{RGB}{228,26,28}
\definecolor{fillColor}{RGB}{228,26,28}

\path[draw=drawColor,line width= 0.3pt,line join=round,line cap=round,fill=fillColor] (116.65,176.34) circle (  1.61);
\definecolor{drawColor}{RGB}{152,78,163}
\definecolor{fillColor}{RGB}{152,78,163}

\path[draw=drawColor,line width= 0.3pt,line join=round,line cap=round,fill=fillColor] (116.65,158.86) circle (  1.61);
\definecolor{drawColor}{RGB}{55,126,184}
\definecolor{fillColor}{RGB}{55,126,184}

\path[draw=drawColor,line width= 0.3pt,line join=round,line cap=round,fill=fillColor] (116.65,106.92) circle (  1.61);
\definecolor{drawColor}{RGB}{77,175,74}
\definecolor{fillColor}{RGB}{77,175,74}

\path[draw=drawColor,line width= 0.3pt,line join=round,line cap=round,fill=fillColor] (133.95,190.60) circle (  1.61);
\definecolor{drawColor}{RGB}{228,26,28}
\definecolor{fillColor}{RGB}{228,26,28}

\path[draw=drawColor,line width= 0.3pt,line join=round,line cap=round,fill=fillColor] (133.95,206.91) circle (  1.61);
\definecolor{drawColor}{RGB}{255,127,0}
\definecolor{fillColor}{RGB}{255,127,0}

\path[draw=drawColor,line width= 0.3pt,line join=round,line cap=round,fill=fillColor] (133.95,200.24) circle (  1.61);
\definecolor{drawColor}{RGB}{152,78,163}
\definecolor{fillColor}{RGB}{152,78,163}

\path[draw=drawColor,line width= 0.3pt,line join=round,line cap=round,fill=fillColor] (133.95,145.91) circle (  1.61);
\definecolor{drawColor}{RGB}{55,126,184}
\definecolor{fillColor}{RGB}{55,126,184}

\path[draw=drawColor,line width= 0.3pt,line join=round,line cap=round,fill=fillColor] (133.95,105.49) circle (  1.61);
\definecolor{drawColor}{RGB}{77,175,74}
\definecolor{fillColor}{RGB}{77,175,74}

\path[draw=drawColor,line width= 0.3pt,line join=round,line cap=round,fill=fillColor] (151.24,184.98) circle (  1.61);
\definecolor{drawColor}{RGB}{255,127,0}
\definecolor{fillColor}{RGB}{255,127,0}

\path[draw=drawColor,line width= 0.3pt,line join=round,line cap=round,fill=fillColor] (151.24,183.40) circle (  1.61);
\definecolor{drawColor}{RGB}{228,26,28}
\definecolor{fillColor}{RGB}{228,26,28}

\path[draw=drawColor,line width= 0.3pt,line join=round,line cap=round,fill=fillColor] (151.24,185.24) circle (  1.61);
\definecolor{drawColor}{RGB}{152,78,163}
\definecolor{fillColor}{RGB}{152,78,163}

\path[draw=drawColor,line width= 0.3pt,line join=round,line cap=round,fill=fillColor] (151.24,121.39) circle (  1.61);
\definecolor{drawColor}{RGB}{55,126,184}
\definecolor{fillColor}{RGB}{55,126,184}

\path[draw=drawColor,line width= 0.3pt,line join=round,line cap=round,fill=fillColor] (151.24,106.66) circle (  1.61);
\definecolor{drawColor}{RGB}{77,175,74}
\definecolor{fillColor}{RGB}{77,175,74}

\path[draw=drawColor,line width= 0.3pt,line join=round,line cap=round,fill=fillColor] (168.53,179.04) circle (  1.61);
\definecolor{drawColor}{RGB}{255,127,0}
\definecolor{fillColor}{RGB}{255,127,0}

\path[draw=drawColor,line width= 0.3pt,line join=round,line cap=round,fill=fillColor] (168.53,196.10) circle (  1.61);
\definecolor{drawColor}{RGB}{228,26,28}
\definecolor{fillColor}{RGB}{228,26,28}

\path[draw=drawColor,line width= 0.3pt,line join=round,line cap=round,fill=fillColor] (168.53,198.44) circle (  1.61);
\definecolor{drawColor}{RGB}{55,126,184}
\definecolor{fillColor}{RGB}{55,126,184}

\path[draw=drawColor,line width= 0.3pt,line join=round,line cap=round,fill=fillColor] (168.53,107.64) circle (  1.61);
\definecolor{drawColor}{RGB}{152,78,163}
\definecolor{fillColor}{RGB}{152,78,163}

\path[draw=drawColor,line width= 0.3pt,line join=round,line cap=round,fill=fillColor] (168.53,113.94) circle (  1.61);
\definecolor{drawColor}{RGB}{77,175,74}
\definecolor{fillColor}{RGB}{77,175,74}

\path[draw=drawColor,line width= 0.3pt,line join=round,line cap=round,fill=fillColor] (185.83,181.59) circle (  1.61);
\definecolor{drawColor}{RGB}{255,127,0}
\definecolor{fillColor}{RGB}{255,127,0}

\path[draw=drawColor,line width= 0.3pt,line join=round,line cap=round,fill=fillColor] (185.83,173.85) circle (  1.61);
\definecolor{drawColor}{RGB}{228,26,28}
\definecolor{fillColor}{RGB}{228,26,28}

\path[draw=drawColor,line width= 0.3pt,line join=round,line cap=round,fill=fillColor] (185.83,174.08) circle (  1.61);
\definecolor{drawColor}{RGB}{152,78,163}
\definecolor{fillColor}{RGB}{152,78,163}

\path[draw=drawColor,line width= 0.3pt,line join=round,line cap=round,fill=fillColor] (185.83,127.65) circle (  1.61);
\definecolor{drawColor}{RGB}{55,126,184}
\definecolor{fillColor}{RGB}{55,126,184}

\path[draw=drawColor,line width= 0.3pt,line join=round,line cap=round,fill=fillColor] (185.83,107.06) circle (  1.61);
\definecolor{drawColor}{RGB}{77,175,74}
\definecolor{fillColor}{RGB}{77,175,74}

\path[draw=drawColor,line width= 0.3pt,line join=round,line cap=round,fill=fillColor] (203.12,179.25) circle (  1.61);
\definecolor{drawColor}{RGB}{255,127,0}
\definecolor{fillColor}{RGB}{255,127,0}

\path[draw=drawColor,line width= 0.3pt,line join=round,line cap=round,fill=fillColor] (203.12,184.99) circle (  1.61);
\definecolor{drawColor}{RGB}{228,26,28}
\definecolor{fillColor}{RGB}{228,26,28}

\path[draw=drawColor,line width= 0.3pt,line join=round,line cap=round,fill=fillColor] (203.12,170.47) circle (  1.61);
\definecolor{drawColor}{RGB}{152,78,163}
\definecolor{fillColor}{RGB}{152,78,163}

\path[draw=drawColor,line width= 0.3pt,line join=round,line cap=round,fill=fillColor] (203.12,125.21) circle (  1.61);
\definecolor{drawColor}{RGB}{55,126,184}
\definecolor{fillColor}{RGB}{55,126,184}

\path[draw=drawColor,line width= 0.3pt,line join=round,line cap=round,fill=fillColor] (203.12,102.55) circle (  1.61);
\definecolor{drawColor}{RGB}{77,175,74}
\definecolor{fillColor}{RGB}{77,175,74}

\path[draw=drawColor,line width= 0.3pt,line join=round,line cap=round,fill=fillColor] (220.41,174.02) circle (  1.61);
\definecolor{drawColor}{RGB}{255,127,0}
\definecolor{fillColor}{RGB}{255,127,0}

\path[draw=drawColor,line width= 0.3pt,line join=round,line cap=round,fill=fillColor] (220.41,181.39) circle (  1.61);
\definecolor{drawColor}{RGB}{228,26,28}
\definecolor{fillColor}{RGB}{228,26,28}

\path[draw=drawColor,line width= 0.3pt,line join=round,line cap=round,fill=fillColor] (220.41,164.87) circle (  1.61);
\definecolor{drawColor}{RGB}{152,78,163}
\definecolor{fillColor}{RGB}{152,78,163}

\path[draw=drawColor,line width= 0.3pt,line join=round,line cap=round,fill=fillColor] (220.41,119.28) circle (  1.61);
\definecolor{drawColor}{RGB}{55,126,184}
\definecolor{fillColor}{RGB}{55,126,184}

\path[draw=drawColor,line width= 0.3pt,line join=round,line cap=round,fill=fillColor] (220.41,101.76) circle (  1.61);
\definecolor{drawColor}{RGB}{255,127,0}
\definecolor{fillColor}{RGB}{255,127,0}

\path[draw=drawColor,line width= 0.3pt,line join=round,line cap=round,fill=fillColor] (237.71,185.93) circle (  1.61);
\definecolor{drawColor}{RGB}{77,175,74}
\definecolor{fillColor}{RGB}{77,175,74}

\path[draw=drawColor,line width= 0.3pt,line join=round,line cap=round,fill=fillColor] (237.71,156.51) circle (  1.61);
\definecolor{drawColor}{RGB}{228,26,28}
\definecolor{fillColor}{RGB}{228,26,28}

\path[draw=drawColor,line width= 0.3pt,line join=round,line cap=round,fill=fillColor] (237.71,150.96) circle (  1.61);
\definecolor{drawColor}{RGB}{152,78,163}
\definecolor{fillColor}{RGB}{152,78,163}

\path[draw=drawColor,line width= 0.3pt,line join=round,line cap=round,fill=fillColor] (237.71,133.85) circle (  1.61);
\definecolor{drawColor}{RGB}{55,126,184}
\definecolor{fillColor}{RGB}{55,126,184}

\path[draw=drawColor,line width= 0.3pt,line join=round,line cap=round,fill=fillColor] (237.71, 97.57) circle (  1.61);
\definecolor{drawColor}{RGB}{255,127,0}
\definecolor{fillColor}{RGB}{255,127,0}

\path[draw=drawColor,line width= 0.3pt,line join=round,line cap=round,fill=fillColor] (255.00,183.13) circle (  1.61);
\definecolor{drawColor}{RGB}{77,175,74}
\definecolor{fillColor}{RGB}{77,175,74}

\path[draw=drawColor,line width= 0.3pt,line join=round,line cap=round,fill=fillColor] (255.00,157.76) circle (  1.61);
\definecolor{drawColor}{RGB}{228,26,28}
\definecolor{fillColor}{RGB}{228,26,28}

\path[draw=drawColor,line width= 0.3pt,line join=round,line cap=round,fill=fillColor] (255.00,153.79) circle (  1.61);
\definecolor{drawColor}{RGB}{152,78,163}
\definecolor{fillColor}{RGB}{152,78,163}

\path[draw=drawColor,line width= 0.3pt,line join=round,line cap=round,fill=fillColor] (255.00,134.65) circle (  1.61);
\definecolor{drawColor}{RGB}{55,126,184}
\definecolor{fillColor}{RGB}{55,126,184}

\path[draw=drawColor,line width= 0.3pt,line join=round,line cap=round,fill=fillColor] (255.00, 97.88) circle (  1.61);
\definecolor{drawColor}{RGB}{255,127,0}
\definecolor{fillColor}{RGB}{255,127,0}

\path[draw=drawColor,line width= 0.3pt,line join=round,line cap=round,fill=fillColor] (272.30,174.18) circle (  1.61);
\definecolor{drawColor}{RGB}{77,175,74}
\definecolor{fillColor}{RGB}{77,175,74}

\path[draw=drawColor,line width= 0.3pt,line join=round,line cap=round,fill=fillColor] (272.30,157.32) circle (  1.61);
\definecolor{drawColor}{RGB}{228,26,28}
\definecolor{fillColor}{RGB}{228,26,28}

\path[draw=drawColor,line width= 0.3pt,line join=round,line cap=round,fill=fillColor] (272.30,150.55) circle (  1.61);
\definecolor{drawColor}{RGB}{152,78,163}
\definecolor{fillColor}{RGB}{152,78,163}

\path[draw=drawColor,line width= 0.3pt,line join=round,line cap=round,fill=fillColor] (272.30,112.55) circle (  1.61);
\definecolor{drawColor}{RGB}{55,126,184}
\definecolor{fillColor}{RGB}{55,126,184}

\path[draw=drawColor,line width= 0.3pt,line join=round,line cap=round,fill=fillColor] (272.30,100.46) circle (  1.61);
\definecolor{drawColor}{RGB}{255,127,0}
\definecolor{fillColor}{RGB}{255,127,0}

\path[draw=drawColor,line width= 0.3pt,line join=round,line cap=round,fill=fillColor] (289.59,171.58) circle (  1.61);
\definecolor{drawColor}{RGB}{77,175,74}
\definecolor{fillColor}{RGB}{77,175,74}

\path[draw=drawColor,line width= 0.3pt,line join=round,line cap=round,fill=fillColor] (289.59,155.83) circle (  1.61);
\definecolor{drawColor}{RGB}{228,26,28}
\definecolor{fillColor}{RGB}{228,26,28}

\path[draw=drawColor,line width= 0.3pt,line join=round,line cap=round,fill=fillColor] (289.59,148.42) circle (  1.61);
\definecolor{drawColor}{RGB}{55,126,184}
\definecolor{fillColor}{RGB}{55,126,184}

\path[draw=drawColor,line width= 0.3pt,line join=round,line cap=round,fill=fillColor] (289.59,102.21) circle (  1.61);
\definecolor{drawColor}{RGB}{152,78,163}
\definecolor{fillColor}{RGB}{152,78,163}

\path[draw=drawColor,line width= 0.3pt,line join=round,line cap=round,fill=fillColor] (289.59,112.71) circle (  1.61);
\definecolor{drawColor}{RGB}{255,127,0}
\definecolor{fillColor}{RGB}{255,127,0}

\path[draw=drawColor,line width= 0.3pt,line join=round,line cap=round,fill=fillColor] (306.88,165.95) circle (  1.61);
\definecolor{drawColor}{RGB}{77,175,74}
\definecolor{fillColor}{RGB}{77,175,74}

\path[draw=drawColor,line width= 0.3pt,line join=round,line cap=round,fill=fillColor] (306.88,152.84) circle (  1.61);
\definecolor{drawColor}{RGB}{228,26,28}
\definecolor{fillColor}{RGB}{228,26,28}

\path[draw=drawColor,line width= 0.3pt,line join=round,line cap=round,fill=fillColor] (306.88,154.72) circle (  1.61);
\definecolor{drawColor}{RGB}{152,78,163}
\definecolor{fillColor}{RGB}{152,78,163}

\path[draw=drawColor,line width= 0.3pt,line join=round,line cap=round,fill=fillColor] (306.88,113.41) circle (  1.61);
\definecolor{drawColor}{RGB}{55,126,184}
\definecolor{fillColor}{RGB}{55,126,184}

\path[draw=drawColor,line width= 0.3pt,line join=round,line cap=round,fill=fillColor] (306.88, 98.29) circle (  1.61);
\definecolor{drawColor}{RGB}{255,127,0}
\definecolor{fillColor}{RGB}{255,127,0}

\path[draw=drawColor,line width= 0.3pt,line join=round,line cap=round,fill=fillColor] (324.18,167.36) circle (  1.61);
\definecolor{drawColor}{RGB}{77,175,74}
\definecolor{fillColor}{RGB}{77,175,74}

\path[draw=drawColor,line width= 0.3pt,line join=round,line cap=round,fill=fillColor] (324.18,154.70) circle (  1.61);
\definecolor{drawColor}{RGB}{228,26,28}
\definecolor{fillColor}{RGB}{228,26,28}

\path[draw=drawColor,line width= 0.3pt,line join=round,line cap=round,fill=fillColor] (324.18,152.43) circle (  1.61);
\definecolor{drawColor}{RGB}{152,78,163}
\definecolor{fillColor}{RGB}{152,78,163}

\path[draw=drawColor,line width= 0.3pt,line join=round,line cap=round,fill=fillColor] (324.18,110.75) circle (  1.61);
\definecolor{drawColor}{RGB}{55,126,184}
\definecolor{fillColor}{RGB}{55,126,184}

\path[draw=drawColor,line width= 0.3pt,line join=round,line cap=round,fill=fillColor] (324.18,102.38) circle (  1.61);
\definecolor{drawColor}{RGB}{255,127,0}
\definecolor{fillColor}{RGB}{255,127,0}

\path[draw=drawColor,line width= 0.3pt,line join=round,line cap=round,fill=fillColor] (341.47,166.58) circle (  1.61);
\definecolor{drawColor}{RGB}{77,175,74}
\definecolor{fillColor}{RGB}{77,175,74}

\path[draw=drawColor,line width= 0.3pt,line join=round,line cap=round,fill=fillColor] (341.47,151.28) circle (  1.61);
\definecolor{drawColor}{RGB}{228,26,28}
\definecolor{fillColor}{RGB}{228,26,28}

\path[draw=drawColor,line width= 0.3pt,line join=round,line cap=round,fill=fillColor] (341.47,153.12) circle (  1.61);
\definecolor{drawColor}{RGB}{55,126,184}
\definecolor{fillColor}{RGB}{55,126,184}

\path[draw=drawColor,line width= 0.3pt,line join=round,line cap=round,fill=fillColor] (341.47,103.71) circle (  1.61);
\definecolor{drawColor}{RGB}{152,78,163}
\definecolor{fillColor}{RGB}{152,78,163}

\path[draw=drawColor,line width= 0.3pt,line join=round,line cap=round,fill=fillColor] (341.47,113.52) circle (  1.61);
\definecolor{drawColor}{RGB}{77,175,74}
\definecolor{fillColor}{RGB}{77,175,74}

\path[draw=drawColor,line width= 0.3pt,line join=round,line cap=round,fill=fillColor] (358.76,151.42) circle (  1.61);
\definecolor{drawColor}{RGB}{255,127,0}
\definecolor{fillColor}{RGB}{255,127,0}

\path[draw=drawColor,line width= 0.3pt,line join=round,line cap=round,fill=fillColor] (358.76,167.64) circle (  1.61);
\definecolor{drawColor}{RGB}{55,126,184}
\definecolor{fillColor}{RGB}{55,126,184}

\path[draw=drawColor,line width= 0.3pt,line join=round,line cap=round,fill=fillColor] (358.76,105.28) circle (  1.61);
\definecolor{drawColor}{RGB}{228,26,28}
\definecolor{fillColor}{RGB}{228,26,28}

\path[draw=drawColor,line width= 0.3pt,line join=round,line cap=round,fill=fillColor] (358.76,153.57) circle (  1.61);
\definecolor{drawColor}{RGB}{152,78,163}
\definecolor{fillColor}{RGB}{152,78,163}

\path[draw=drawColor,line width= 0.3pt,line join=round,line cap=round,fill=fillColor] (358.76,107.09) circle (  1.61);
\definecolor{drawColor}{RGB}{255,127,0}
\definecolor{fillColor}{RGB}{255,127,0}

\path[draw=drawColor,line width= 0.3pt,line join=round,line cap=round,fill=fillColor] (376.06,166.30) circle (  1.61);
\definecolor{drawColor}{RGB}{77,175,74}
\definecolor{fillColor}{RGB}{77,175,74}

\path[draw=drawColor,line width= 0.3pt,line join=round,line cap=round,fill=fillColor] (376.06,153.31) circle (  1.61);
\definecolor{drawColor}{RGB}{228,26,28}
\definecolor{fillColor}{RGB}{228,26,28}

\path[draw=drawColor,line width= 0.3pt,line join=round,line cap=round,fill=fillColor] (376.06,159.58) circle (  1.61);
\definecolor{drawColor}{RGB}{55,126,184}
\definecolor{fillColor}{RGB}{55,126,184}

\path[draw=drawColor,line width= 0.3pt,line join=round,line cap=round,fill=fillColor] (376.06,114.24) circle (  1.61);
\definecolor{drawColor}{RGB}{152,78,163}
\definecolor{fillColor}{RGB}{152,78,163}

\path[draw=drawColor,line width= 0.3pt,line join=round,line cap=round,fill=fillColor] (376.06,105.80) circle (  1.61);
\definecolor{drawColor}{gray}{0.50}

\path[draw=drawColor,line width= 0.5pt,dash pattern=on 4pt off 4pt ,line join=round] (220.41, 75.66) -- (220.41,247.95);
\end{scope}
\begin{scope}
\path[clip] (  0.00,  0.00) rectangle (397.48,252.94);
\definecolor{drawColor}{gray}{0.30}

\node[text=drawColor,anchor=base east,inner sep=0pt, outer sep=0pt, scale=  0.80] at ( 26.55, 80.73) {0.0};

\node[text=drawColor,anchor=base east,inner sep=0pt, outer sep=0pt, scale=  0.80] at ( 26.55,103.11) {0.1};

\node[text=drawColor,anchor=base east,inner sep=0pt, outer sep=0pt, scale=  0.80] at ( 26.55,125.48) {0.2};

\node[text=drawColor,anchor=base east,inner sep=0pt, outer sep=0pt, scale=  0.80] at ( 26.55,147.86) {0.3};

\node[text=drawColor,anchor=base east,inner sep=0pt, outer sep=0pt, scale=  0.80] at ( 26.55,170.23) {0.4};

\node[text=drawColor,anchor=base east,inner sep=0pt, outer sep=0pt, scale=  0.80] at ( 26.55,192.61) {0.5};

\node[text=drawColor,anchor=base east,inner sep=0pt, outer sep=0pt, scale=  0.80] at ( 26.55,214.98) {0.6};

\node[text=drawColor,anchor=base east,inner sep=0pt, outer sep=0pt, scale=  0.80] at ( 26.55,237.36) {0.7};
\end{scope}
\begin{scope}
\path[clip] (  0.00,  0.00) rectangle (397.48,252.94);
\definecolor{drawColor}{gray}{0.30}

\node[text=drawColor,anchor=base,inner sep=0pt, outer sep=0pt, scale=  0.80] at ( 64.77, 65.65) {2008};

\node[text=drawColor,anchor=base,inner sep=0pt, outer sep=0pt, scale=  0.80] at ( 99.36, 65.65) {2010};

\node[text=drawColor,anchor=base,inner sep=0pt, outer sep=0pt, scale=  0.80] at (133.95, 65.65) {2012};

\node[text=drawColor,anchor=base,inner sep=0pt, outer sep=0pt, scale=  0.80] at (168.53, 65.65) {2014};

\node[text=drawColor,anchor=base,inner sep=0pt, outer sep=0pt, scale=  0.80] at (203.12, 65.65) {2016};

\node[text=drawColor,anchor=base,inner sep=0pt, outer sep=0pt, scale=  0.80] at (237.71, 65.65) {2018};

\node[text=drawColor,anchor=base,inner sep=0pt, outer sep=0pt, scale=  0.80] at (272.30, 65.65) {2020};

\node[text=drawColor,anchor=base,inner sep=0pt, outer sep=0pt, scale=  0.80] at (306.88, 65.65) {2022};

\node[text=drawColor,anchor=base,inner sep=0pt, outer sep=0pt, scale=  0.80] at (341.47, 65.65) {2024};

\node[text=drawColor,anchor=base,inner sep=0pt, outer sep=0pt, scale=  0.80] at (376.06, 65.65) {2026};
\end{scope}
\begin{scope}
\path[clip] (  0.00,  0.00) rectangle (397.48,252.94);
\definecolor{drawColor}{RGB}{0,0,0}

\node[text=drawColor,anchor=base,inner sep=0pt, outer sep=0pt, scale=  1.00] at (211.77, 54.71) {Year};
\end{scope}
\begin{scope}
\path[clip] (  0.00,  0.00) rectangle (397.48,252.94);
\definecolor{drawColor}{RGB}{0,0,0}

\node[text=drawColor,rotate= 90.00,anchor=base,inner sep=0pt, outer sep=0pt, scale=  1.00] at ( 11.89,161.80) {MRR (Adamic-Adar)};
\end{scope}
\begin{scope}
\path[clip] (  0.00,  0.00) rectangle (397.48,252.94);
\definecolor{drawColor}{RGB}{228,26,28}

\path[draw=drawColor,line width= 0.9pt,line join=round] ( 55.42, 32.07) -- ( 64.53, 32.07);
\definecolor{fillColor}{RGB}{228,26,28}

\path[draw=drawColor,line width= 0.3pt,line join=round,line cap=round,fill=fillColor] ( 59.97, 32.07) circle (  1.61);
\end{scope}
\begin{scope}
\path[clip] (  0.00,  0.00) rectangle (397.48,252.94);
\definecolor{drawColor}{RGB}{55,126,184}

\path[draw=drawColor,line width= 0.9pt,line join=round] ( 55.42, 15.69) -- ( 64.53, 15.69);
\definecolor{fillColor}{RGB}{55,126,184}

\path[draw=drawColor,line width= 0.3pt,line join=round,line cap=round,fill=fillColor] ( 59.97, 15.69) circle (  1.61);
\end{scope}
\begin{scope}
\path[clip] (  0.00,  0.00) rectangle (397.48,252.94);
\definecolor{drawColor}{RGB}{77,175,74}

\path[draw=drawColor,line width= 0.9pt,line join=round] (150.73, 32.07) -- (159.84, 32.07);
\definecolor{fillColor}{RGB}{77,175,74}

\path[draw=drawColor,line width= 0.3pt,line join=round,line cap=round,fill=fillColor] (155.28, 32.07) circle (  1.61);
\end{scope}
\begin{scope}
\path[clip] (  0.00,  0.00) rectangle (397.48,252.94);
\definecolor{drawColor}{RGB}{152,78,163}

\path[draw=drawColor,line width= 0.9pt,line join=round] (150.73, 15.69) -- (159.84, 15.69);
\definecolor{fillColor}{RGB}{152,78,163}

\path[draw=drawColor,line width= 0.3pt,line join=round,line cap=round,fill=fillColor] (155.28, 15.69) circle (  1.61);
\end{scope}
\begin{scope}
\path[clip] (  0.00,  0.00) rectangle (397.48,252.94);
\definecolor{drawColor}{RGB}{255,127,0}

\path[draw=drawColor,line width= 0.9pt,line join=round] (256.13, 32.07) -- (265.24, 32.07);
\definecolor{fillColor}{RGB}{255,127,0}

\path[draw=drawColor,line width= 0.3pt,line join=round,line cap=round,fill=fillColor] (260.68, 32.07) circle (  1.61);
\end{scope}
\begin{scope}
\path[clip] (  0.00,  0.00) rectangle (397.48,252.94);
\definecolor{drawColor}{RGB}{0,0,0}

\node[text=drawColor,anchor=base west,inner sep=0pt, outer sep=0pt, scale=  0.70] at ( 70.66, 29.66) {Admin (Administrative)};
\end{scope}
\begin{scope}
\path[clip] (  0.00,  0.00) rectangle (397.48,252.94);
\definecolor{drawColor}{RGB}{0,0,0}

\node[text=drawColor,anchor=base west,inner sep=0pt, outer sep=0pt, scale=  0.70] at ( 70.66, 13.28) {CivCode (Civil Code)};
\end{scope}
\begin{scope}
\path[clip] (  0.00,  0.00) rectangle (397.48,252.94);
\definecolor{drawColor}{RGB}{0,0,0}

\node[text=drawColor,anchor=base west,inner sep=0pt, outer sep=0pt, scale=  0.70] at (165.97, 29.66) {CivProc (Civil Procedure)};
\end{scope}
\begin{scope}
\path[clip] (  0.00,  0.00) rectangle (397.48,252.94);
\definecolor{drawColor}{RGB}{0,0,0}

\node[text=drawColor,anchor=base west,inner sep=0pt, outer sep=0pt, scale=  0.70] at (165.97, 13.28) {CrimCode (Criminal Code)};
\end{scope}
\begin{scope}
\path[clip] (  0.00,  0.00) rectangle (397.48,252.94);
\definecolor{drawColor}{RGB}{0,0,0}

\node[text=drawColor,anchor=base west,inner sep=0pt, outer sep=0pt, scale=  0.70] at (271.37, 29.66) {CrimProc (Criminal Procedure)};
\end{scope}
\end{tikzpicture}

%% file: figures/fig7_embedding_drift.tex
\begin{tikzpicture}[x=1pt,y=1pt]
\definecolor{fillColor}{RGB}{255,255,255}
\path[use as bounding box,fill=fillColor,fill opacity=0.00] (0,0) rectangle (397.48,216.81);
\begin{scope}
\path[clip] (  0.00,  0.00) rectangle (397.48,216.81);
\definecolor{fillColor}{RGB}{255,255,255}

\path[fill=fillColor] (  0.00,  0.00) rectangle (397.48,216.81);
\end{scope}
\begin{scope}
\path[clip] ( 35.05, 31.03) rectangle (392.48,211.81);
\definecolor{drawColor}{gray}{0.92}

\path[draw=drawColor,line width= 0.5pt,line join=round] ( 35.05, 39.25) --
	(392.48, 39.25);

\path[draw=drawColor,line width= 0.5pt,line join=round] ( 35.05, 72.12) --
	(392.48, 72.12);

\path[draw=drawColor,line width= 0.5pt,line join=round] ( 35.05,104.98) --
	(392.48,104.98);

\path[draw=drawColor,line width= 0.5pt,line join=round] ( 35.05,137.85) --
	(392.48,137.85);

\path[draw=drawColor,line width= 0.5pt,line join=round] ( 35.05,170.72) --
	(392.48,170.72);

\path[draw=drawColor,line width= 0.5pt,line join=round] ( 35.05,203.59) --
	(392.48,203.59);

\path[draw=drawColor,line width= 0.5pt,line join=round] (102.07, 31.03) --
	(102.07,211.81);

\path[draw=drawColor,line width= 0.5pt,line join=round] (213.77, 31.03) --
	(213.77,211.81);

\path[draw=drawColor,line width= 0.5pt,line join=round] (325.47, 31.03) --
	(325.47,211.81);
\definecolor{drawColor}{gray}{0.20}
\definecolor{fillColor}{gray}{0.20}

\path[draw=drawColor,line width= 0.4pt,line join=round,line cap=round,fill=fillColor] (102.07,173.68) circle (  1.21);

\path[draw=drawColor,line width= 0.4pt,line join=round,line cap=round,fill=fillColor] (102.07,139.66) circle (  1.21);

\path[draw=drawColor,line width= 0.4pt,line join=round,line cap=round,fill=fillColor] (102.07,143.77) circle (  1.21);

\path[draw=drawColor,line width= 0.4pt,line join=round,line cap=round,fill=fillColor] (102.07,163.99) circle (  1.21);

\path[draw=drawColor,line width= 0.5pt,line join=round] (102.07,118.50) -- (102.07,138.84);

\path[draw=drawColor,line width= 0.5pt,line join=round] (102.07,104.41) -- (102.07, 93.65);
\definecolor{fillColor}{RGB}{102,194,165}

\path[draw=drawColor,line width= 0.5pt,fill=fillColor] ( 68.56,118.50) --
	( 68.56,104.41) --
	(135.58,104.41) --
	(135.58,118.50) --
	( 68.56,118.50) --
	cycle;

\path[draw=drawColor,line width= 1.0pt] ( 68.56,112.30) -- (135.58,112.30);
\definecolor{fillColor}{gray}{0.20}

\path[draw=drawColor,line width= 0.4pt,line join=round,line cap=round,fill=fillColor] (325.47,132.60) circle (  1.21);

\path[draw=drawColor,line width= 0.4pt,line join=round,line cap=round,fill=fillColor] (325.47,135.72) circle (  1.21);

\path[draw=drawColor,line width= 0.5pt,line join=round] (325.47,106.63) -- (325.47,115.01);

\path[draw=drawColor,line width= 0.5pt,line join=round] (325.47, 96.93) -- (325.47, 86.91);
\definecolor{fillColor}{RGB}{252,141,98}

\path[draw=drawColor,line width= 0.5pt,fill=fillColor] (291.96,106.63) --
	(291.96, 96.93) --
	(358.98, 96.93) --
	(358.98,106.63) --
	(291.96,106.63) --
	cycle;

\path[draw=drawColor,line width= 1.0pt] (291.96,102.19) -- (358.98,102.19);

\path[draw=drawColor,line width= 0.5pt,line join=round] (213.77,114.52) -- (213.77,128.65);

\path[draw=drawColor,line width= 0.5pt,line join=round] (213.77,100.55) -- (213.77, 94.80);
\definecolor{fillColor}{RGB}{141,160,203}

\path[draw=drawColor,line width= 0.5pt,fill=fillColor] (180.26,114.52) --
	(180.26,100.55) --
	(247.28,100.55) --
	(247.28,114.52) --
	(180.26,114.52) --
	cycle;

\path[draw=drawColor,line width= 1.0pt] (180.26,104.16) -- (247.28,104.16);
\end{scope}
\begin{scope}
\path[clip] (  0.00,  0.00) rectangle (397.48,216.81);
\definecolor{drawColor}{gray}{0.30}

\node[text=drawColor,anchor=base east,inner sep=0pt, outer sep=0pt, scale=  0.80] at ( 30.55, 36.49) {0.00};

\node[text=drawColor,anchor=base east,inner sep=0pt, outer sep=0pt, scale=  0.80] at ( 30.55, 69.36) {0.02};

\node[text=drawColor,anchor=base east,inner sep=0pt, outer sep=0pt, scale=  0.80] at ( 30.55,102.23) {0.04};

\node[text=drawColor,anchor=base east,inner sep=0pt, outer sep=0pt, scale=  0.80] at ( 30.55,135.10) {0.06};

\node[text=drawColor,anchor=base east,inner sep=0pt, outer sep=0pt, scale=  0.80] at ( 30.55,167.97) {0.08};

\node[text=drawColor,anchor=base east,inner sep=0pt, outer sep=0pt, scale=  0.80] at ( 30.55,200.84) {0.10};
\end{scope}
\begin{scope}
\path[clip] (  0.00,  0.00) rectangle (397.48,216.81);
\definecolor{drawColor}{gray}{0.30}

\node[text=drawColor,rotate= 25.00,anchor=base east,inner sep=0pt, outer sep=0pt, scale=  0.80] at (104.40, 21.54) {CivProc};

\node[text=drawColor,rotate= 25.00,anchor=base east,inner sep=0pt, outer sep=0pt, scale=  0.80] at (216.10, 21.54) {CrimProc};

\node[text=drawColor,rotate= 25.00,anchor=base east,inner sep=0pt, outer sep=0pt, scale=  0.80] at (327.79, 21.54) {CrimCode};
\end{scope}
\begin{scope}
\path[clip] (  0.00,  0.00) rectangle (397.48,216.81);
\definecolor{drawColor}{RGB}{0,0,0}

\node[text=drawColor,rotate= 90.00,anchor=base,inner sep=0pt, outer sep=0pt, scale=  1.00] at ( 11.89,121.42) {Embedding drift (2012 $\to$ 2024)};
\end{scope}
\end{tikzpicture}